\documentclass[10pt,journal,compsoc]{IEEEtran}
\usepackage{url}            
\usepackage{booktabs}       
\usepackage{amsfonts}       
\usepackage{nicefrac}       
\usepackage{microtype}      
\usepackage{graphicx}

\usepackage{multirow}
\usepackage{comment}
\usepackage{amsmath,amssymb} 
\usepackage{color}
\usepackage{array}
\usepackage{pifont}
\usepackage{algorithmic}
\usepackage{algorithm}
\usepackage{footmisc}
\usepackage[colorlinks,linkcolor=red]{hyperref}
\usepackage[caption=false]{subfig}

\ifCLASSOPTIONcompsoc
  \usepackage[nocompress]{cite}
\else
  \usepackage{cite}
\fi
\ifCLASSINFOpdf
\else
\fi
\begin{document}
%
\title{End2End Occluded Face Recognition by Masking Corrupted Features}

\author{Haibo~Qiu, Dihong~Gong, Zhifeng~Li,~\IEEEmembership{Senior Member,~IEEE}, \\ Wei~Liu,~\IEEEmembership{Senior Member,~IEEE}, 
        and Dacheng~Tao,~\IEEEmembership{Fellow,~IEEE}
\IEEEcompsocitemizethanks{
    \IEEEcompsocthanksitem Haibo Qiu and Dacheng Tao are with the JD Explore Academy, China (E-mail: \{qiuhaibo1, taodacheng\}@jd.com).  
	\IEEEcompsocthanksitem Dihong Gong, Zhifeng Li and Wei Liu are with the Tencent Data Platform, Shenzhen, China. (E-mail: gongdihong@gmail.com; michaelzfli@tencent.com; wl2223@columbia.edu).
	\IEEEcompsocthanksitem Corresponding authors: Zhifeng Li, Wei Liu and Dacheng Tao.
	}
}

\markboth{IEEE Transactions on Pattern Analysis and Machine Intelligence}%
{Shell \MakeLowercase{\textit{et al.}}: Bare Demo of IEEEtran.cls for Computer Society Journals}
%

\IEEEtitleabstractindextext{%
\begin{abstract}
With the recent advancement of deep convolutional neural networks, significant progress has been made in general face recognition. However, the state-of-the-art general face recognition models do not generalize well to occluded face images, which are exactly the common cases in real-world scenarios. The potential reasons are the absences of large-scale occluded face data for training and specific designs for tackling corrupted features brought by occlusions. This paper presents a novel face recognition method that is robust to occlusions based on a single end-to-end deep neural network. Our approach, named FROM (Face Recognition with Occlusion Masks), learns to discover the corrupted features from the deep convolutional neural networks, and clean them by the dynamically learned masks. In addition, we construct massive occluded face images to train FROM effectively and efficiently. FROM is simple yet powerful compared to the existing methods that either rely on external detectors to discover the occlusions or employ shallow models which are less discriminative. Experimental results on the LFW, Megaface challenge 1, RMF2, AR dataset and other simulated occluded/masked datasets confirm that FROM dramatically improves the accuracy under occlusions, and generalizes well on general face recognition. Code is available at https://github.com/haibo-qiu/FROM
\end{abstract}

}

\maketitle

\IEEEdisplaynontitleabstractindextext

\IEEEpeerreviewmaketitle

\IEEEraisesectionheading{\section{Introduction}
\label{sec:introduction}}

\IEEEPARstart{F}{ace} recognition has achieved remarkable progress in the past few years, mostly attributing to three factors: 1) sophisticated formulation of loss functions~\cite{deng2019arcface,hoffer2015deep,liu2017sphereface,wang2018cosface,wen2016discriminative}, 2) carefully designed convolutional neural network architectures (CNNs)~\cite{he2016deep,huang2017densely,krizhevsky2017imagenet,simonyan2014very,tan2019efficientnet}, and 3) large-scale training datasets~\cite{guo2016ms,huang2008labeled,kemelmacher2016megaface}. However, when applying those face recognition models in real-world scenarios, there are still many challenges to be tackled, \eg, extreme illumination, rare head pose, low resolutions, and occlusions. Without carefully designed architectures and large-scale training datasets, the accuracy of state-of-the-art general face recognition models~\cite{deng2019arcface,liu2017sphereface,wang2018cosface} is significantly degraded under those circumstances as observed by Mehdipour \etal~\cite{mehdipour2016comprehensive}. In this paper, we focus on developing a novel CNNs-based face recognition model to remarkably boost the accuracy in the context of occlusions, while sustaining the competitive performance on general face recognition. 

\begin{figure}[!htb]
\centering
\includegraphics[width=0.85\linewidth]{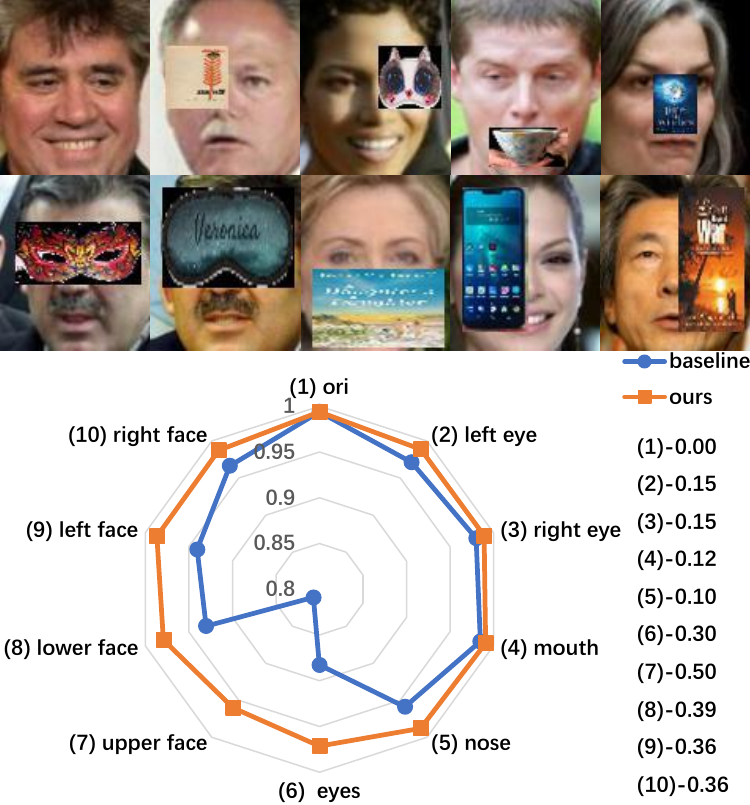}
\caption{The upper part visualizes nine different occlusions on face images. The lower part demonstrates the accuracy comparison between baseline and our method. Note that the right legends indicate the occluded degrees of different occlusions (\eg, (4)-0.12 represents that the area of blocking mouth is $12\%$ of the total face image area).}
\label{fig:diff}
\vspace{-4mm}
\end{figure}

To demonstrate the performance degradation brought by occlusions, we conduct a toy experiment as illustrated by Figure~\ref{fig:diff}. Based on LFW~\cite{huang2008labeled}, we construct nine different occluded LFW datasets by blocking out left eye, right eye, mouth, nose, eyes, upper face, lower face, left face and right face areas separately with random occluders illustrated in Figure~\ref{fig:occluder}. The baseline model is trained on CASIA-WebFace~\cite{yi2014learning} with the LResNet50E-IR architecture~\cite{deng2019arcface} and CosFace loss~\cite{wang2018cosface}. From Figure~\ref{fig:diff}, the baseline achieves an excellent $99.38\%$ accuracy on original LFW, while degrading significantly under occlusions. It only obtains $81.12\%$, $88.28\%$ under the occlusions of upper face and eyes, suggesting that the eyes contain crucial identity information. In contrast, the occlusion of mouth decreases the accuracy slightly, indicating its negligible role during the recognition process. Obviously, our method (orange square) dramatically surpasses the baseline (blue dot) under all occlusions. In particular, the accuracy gap becomes larger when occlusion becomes harder (\eg, $96.03\%$ \vs~$81.12\%$ under the upper face occlusion). Importantly, our method also sustains the competitive accuracy (\ie, $99.38\%$) on the original LFW dataset, which will be detailed in Sec.~\ref{sec:method}. 

To mitigate the degradation caused by occlusions, two different branches of approaches are proposed: Recovering and Removing. Recovering means recovering the occluded facial parts first, and then performing recognition on the recovered face images. Deep learning method is first introduced to this branch by Zhao \etal~\cite{zhao2017robust}. They proposed an LSTM-Autoencoder to restore occluded facial areas in the wild and carried out recognition on the recovered face images. However, this branch of methods usually suffers from the challenge of recovering facial parts while preserving identity information. In contrast, Removing represents first removing the features that are corrupted by occlusions, and then utilizing the remaining clean features for recognition. The key is to find the correspondence between the occlusions of input face images and deep features, which is not trivial because of the complexity of deep neural models. Song \etal \cite{song2019occlusion} divided a face into $K \times K$ blocks and trained $K^2$ deep face models to discover the correspondence between deep features and certain blocks. The obtained correspondences are stored in a dictionary, which is consulted to get the matched masks to clean the corrupted features when testing. Nevertheless, they had to rely on an external occlusion detector to detect the blocked areas, and took massive time to train $K^2$ deep face models. 

In this paper, we propose a simple yet effective method to clean the corrupted features from the deep neural networks with feature masks for robust occluded face recognition. The proposed FROM (Face Recognition with Occlusion Masks) adopts a sub-network (\ie, \textbf{Mask Decoder}) to dynamically decode the accurate feature masks which reveal the locations of corrupted features caused by the occlusion from input face image, and cleans them by the multiplication of features and learned masks. In addition, we propose to utilize the proximity characteristic of occlusions (\eg, the adjacent face regions usually have the same occlusion status) as the supervision for the feature masks learning. Unlike~\cite{song2019occlusion}, we do not need to train multiple models to build a lookup dictionary because FROM can dynamically generate the feature masks according to the input face images. It also makes FROM generalize well to clean face images, and not rely on external occlusion detectors, significantly reducing the inference latency at testing. The contributions of this paper are summarized as follows.

\begin{enumerate}
    \item We introduce FROM, a single-network occluded face recognition method, which can be trained end-to-end to simultaneously learn the feature masks and deep occlusion-robust features. FROM has nearly $N \times$ less FLOPS and $N \times$ less parameters ($N=K^2, K=3$ in~\cite{song2019occlusion}, referring to Table~\ref{table:flops}), while performing significantly more accurately than the most recent state-of-the-art method~\cite{song2019occlusion}.
    \item We propose to learn an ``occlusion to feature mask" mapping with a \textbf{Mask Decoder}, which takes the deep pyramid features as input and captures both local and global information to learn the accurate masks. Besides, the following \textbf{Occlusion Pattern Predictor} guides the \textbf{Mask Decoder} to produce masks which are consistent with occlusion patterns.
    \item Experiments on LFW, Megaface Challenge 1, RMF2, AR Face Dataset and other simulated occluded/masked datasets demonstrate that our approach accomplishes superior accuracy for occluded face recognition, and generalizes very well on general face recognition.
\end{enumerate}

\begin{table}[!hbt]
\center
\vspace{-5mm}
\caption{Comparison between PDSN~\cite{song2019occlusion} and FROM in terms of FLOPs and Parameters. ($\times 9$) indicates that PDSN needs to train nine separated models for nine different occlusion blocks.}
\label{table:flops}
\begin{tabular}{p{2cm} p{2cm}<{\centering} p{2cm}<{\centering}}
\hline
\noalign{\smallskip}
Method  & FLOPs (G)  \quad & Params (M) \quad \\
\noalign{\smallskip}
\hline
\noalign{\smallskip}
PDSN~\cite{song2019occlusion} & 11.08 $\times 9$ & 44.16  $\times 9$ \\
\noalign{\smallskip}
FROM & 12.34 & 52.24 \\
\hline
\end{tabular}
\vspace{-3mm}
\end{table}
\section{Related Work}
\label{sec:related_work}

In this section, the recent advancements in general face recognition are firstly reviewed. Then we explore the development of occluded face recognition and discuss how they differ from our work.

\vspace*{3mm}
\noindent
\textbf{General Face Recognition.} The recent progress in general face recognition is mainly on the improved designs of training loss functions. Their core idea is to enhance the model's discriminative power by maximizing inter-class variance and minimizing intra-class variance. In general, they can be divide into two groups: 1) optimization based on pair-wise distance comparison~\cite{chopra2005learning,hadsell2006dimensionality,hoffer2015deep}, and 2) classification learning with softmax or its variants~\cite{wen2016discriminative,liu2016large,liu2017sphereface,wang2018cosface,deng2019arcface}. The first group of methods usually calculates pair-wise Euclidean distance first, and then obtains the loss to optimize the model according to the relationships of input samples, \eg, Contrastive loss~\cite{chopra2005learning,hadsell2006dimensionality} and Triplet loss~\cite{hoffer2015deep}. They can be employed to effectively train the model when the data is scarce. In contrast, the second group formulates the model training as a classification task learning, which can utilize the potential of large-scale data to enhance the discriminative power of trained models. Center loss~\cite{wen2016discriminative} learns the centers for deep features of each identity, which are then leveraged to reduce intra-class variance. Large margin softmax (L-Softmax)~\cite{liu2016large} improves feature discrimination by adding angular constraints to each identity. Angular softmax (SphereFace)~\cite{liu2017sphereface} improves L-Softmax by normalizing the weights to introduce the hypersphere space. CosFace~\cite{wang2018cosface} proposes an additive margin in the cosine space to enlarge the inter-class variance. ArcFace~\cite{deng2019arcface} introduces an additive angular margin into the angular space and gives a more clear geometric interpretation. UniformFace~\cite{duan2019uniformface} enforces the class centers to uniformly distribute on the feature space to maximize the inter-class distance. In this paper, we adopt the large margin cosine loss~\cite{wang2018cosface} as our training face recognition loss function. 

\begin{figure*}[htbp]
	\centering
	\includegraphics[width=.9\textwidth]{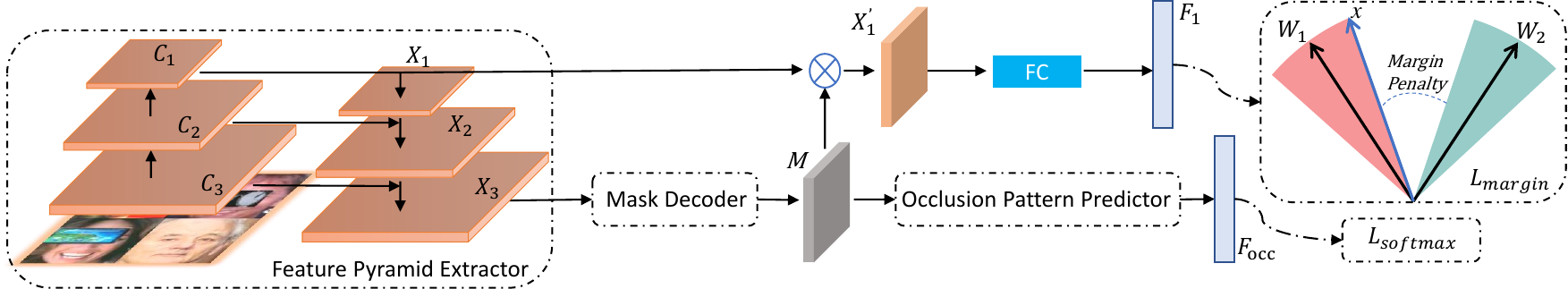}
	\caption{The pipeline of the proposed FROM. FROM first takes a mini-batch which consists of different random occluded and occlusion-free (not paired) face images as input, and generates a feature pyramid (including $X_1, X_2, X_3$). Then $X_3$ is used to decode the masks, which are later applied to $X_1$ to mask out the corrupted feature elements for the final recognition. We also propose to leverage the occlusion patterns as the extra supervision to guide the feature masks learning. The whole network is trained end-to-end.}
	\label{fig:framework}
	\vspace{-3mm}
\end{figure*}

\vspace*{3mm}
\noindent
\textbf{Occluded Face Recognition.} As illustrated by Figure~\ref{fig:diff}, the performance of state-of-the-art general face recognition models is significantly degraded by the occlusions. To alleviate the degradation, current methods for occluded face recognition \cite{wright2008robust,zhou2009face,zhao2017robust,li2005nonparametric,wan2017occlusion,song2019occlusion} are usually the variants of two sets of approaches: 1) recovering occluded facial parts \cite{wright2008robust,yang2011robust,he2011regularized,zhou2009face,li2013structured,zhao2017robust}, and 2) removing features that are corrupted by occlusions. Among the first type of methods, the pioneering work is a sparse representation-based classification (SRC) \cite{wright2008robust} which restores the occluded facial parts using a linear combination of training images. Later, SRC is then improved by either designing a distribution of the sparse constraint term \cite{yang2011robust,he2011regularized}, or characterizing the structure information \cite{zhou2009face,li2013structured}. Deep learning is first introduced to restore occluded faces for recognition by Zhao \etal~\cite{zhao2017robust}. They presented an LSTM-Autoencoder to restore occluded facial areas in the wild and made the recognition on the recovered face images. However, the first type of methods usually performs not quite well as recovering facial parts while preserving strong identity information is still very challenging.

Early studies of the second type are based on shallow models with hand-crafted features such as HOG, SIFT, LBP, etc. Such features have a clear spatial mapping from the input image space to the feature space. As a result, removing corrupted features is straightforward when occlusions are provided \cite{li2005nonparametric,min2011improving,oh2008occlusion}. However, the accuracy of these methods is limited by the shallow architectures. Recent studies have harnessed the unrivalled power of deep models by designing sophisticated algorithms to remove the corrupted deep features. Due to the opacity of the spatial mapping between the input image and the deep features caused by deep CNNs, it is hard to identify the corrupted features even if the locations of occlusions in the input image are provided. Wan \etal~\cite{wan2017occlusion} tackled the problem by adding a MaskNet branch to the middle layer of CNN models, which is expected to assign lower weights to hidden units corrupted by the occlusions. But the middle part of the network contains too much unrelated information and no extra supervisions are given for the learning guidance, so it is hard to reliably identify the corrupted units. The most recent state-of-the-art method, proposed by Song \etal~\cite{song2019occlusion}, cleans the corrupted features with binary masks from the higher level of the network where features are more discriminative. Face image is divided into $K \times K$ blocks, and a dictionary of $K^2$ entries is learned to map each occlusion block to the corresponding feature mask. At the testing stage, it first detects the occluded blocks, and then retrieves the corresponding binary masks to apply on the testing features. This two-stage method has to rely on an external occlusion detector which is another heavy deep network. Additionally, $K^2$ deep face models have to be trained separately at the training stage to learn the dictionary, which makes it inefficient and time-consuming for training.

Instead, FROM contains a \textbf{Mask Decoder} which dynamically predicts the correct feature masks which reveal the locations of corrupted features caused by the occlusion of the input face image, and cleans the deep features with the learned masks for the later recognition. Moreover, our FROM follows the end-to-end paradigm and does not rely on any external off-the-shelf detector. Importantly, comparing to~\cite{song2019occlusion}, we only need to train a single model.
\section{Method}
\label{sec:method}

The proposed FROM, illustrated by Figure~\ref{fig:framework}, is a novel single-network end-to-end method. It takes a mini-batch of randomly occluded and occlusion-free (unpaired) face images as input, and generates pyramid features which are then utilized to decode feature masks. Then, the obtained masks clean the deep features by masking out the corrupted elements via multiplication for the final recognition. The core idea of FROM is to learn accurate feature masks to effectively clean the corrupted features. To implement this idea, we add occlusion information supervision of input face images to the decoded masks learning, which connects the decoded masks and the input occlusion patterns. Specifically, we design a special form of occlusion patterns supervision that considers the proximity characteristic for occlusions, making the learning process of the \textbf{Mask Decoder} more stable and obtaining more accurate masks to yield better accuracy.

In the following, we first discuss the recent deep general face recognition methods using margin-based softmax loss functions. The \textbf{Feature Pyramid Extractor} that extracts the features for masks generation and discriminative recognition is then depicted. Next we explore the details of the \textbf{Mask Decoder} and the learned feature masks. After that, we show the design thinking of the \textbf{Occlusion Pattern Predictor} and the proximity-considered form of occlusion patterns. Lastly, the overall training objective of the entire framework for end-to-end learning is provided.

\subsection{Deep General Face Recognition}
\label{sec:loss}
Recently, margin-based softmax loss functions~\cite{deng2019arcface,liu2017sphereface,wang2018cosface} are widely adopted to significantly boost the face recognition performance, which explicitly introduce the margin penalty in the normalized hypersphere to simultaneously enhance the intra-class compactness and the inter-class discrepancy. In addition, they can be trivially implemented by reformulating the original softmax loss function. Following ~\cite{deng2019arcface}, we adopt a unified formulation for margin-based softmax loss functions as follows:
\begin{equation}
    \begin{split}
        & \mathcal{L}_{margin} =-\frac{1}{N}\sum_{i=1}^{N}\log\frac{e^{s \cdot \delta }}{e^{s \cdot \delta }+\sum_{j\neq  y_i}^{n}e^{s\cos\theta_{j}}},
        \\
        \\
        & \wrt \quad \delta = \cos(m_1\theta_{y_i}+m_2) - m_3,
    \end{split}
    \label{eq:margin}
\end{equation}
where $N$ is the number of training samples, $\theta_j$ indicates the angle between the weight $W_j$ and the feature $x_i$, $y_i$ represents the ground-truth class, $m_{1,2,3}$ are margins, and $s$ is the scale factor. With the formulation in Eq.~\eqref{eq:margin}, different margin-based loss functions can be defined using different configurations of the triplet coefficients $(m_1, m_2, m_3)$. Specifically, for SphereFace~\cite{liu2017sphereface}, ArcFace~\cite{deng2019arcface} and CosFace~\cite{wang2018cosface}, we have $(m_1, 0, 0)$, $(0, m_2, 0)$, and $(0, 0, m_3)$, respectively. 

\subsection{Feature Pyramid Extractor}
\label{sec:extractor}
To simultaneously obtain the spatial-aware features for accurate mask learning and discriminative features for recognition, we adopt the \textbf{Feature Pyramid Extractor}~\cite{lin2017feature} as our backbone network, which has the top-down architecture with lateral connections to build pyramid features. Similar to~\cite{song2019occlusion}, we employ the refined ResNet50\cite{he2016deep} architecture named LResnet50E-IR~\cite{deng2019arcface} as the body of the backbone network. As illustrated in Figure~\ref{fig:framework}, it takes the aligned mixed face images as input and outputs pyramid features $X_1, X_2$ and $X_3$. Note that $X_1$ is the deep discriminative feature to be cleaned, with the size of $N \times C \times H \times W$, where $N$ is the batch size, $C$ is the number of channels, and $H \times W$ are the height and width. $X_3$, containing both the local and global information, is fed into the \textbf{Mask Decoder} to decode the corresponding feature mask $M$ for removing the corrupted elements of $X_1$.

\subsection{Mask Decoder}
\label{sec:mask_decoder}
\begin{figure}[t]
    \centering
    \subfloat[MD]{\includegraphics[width=.45\linewidth]{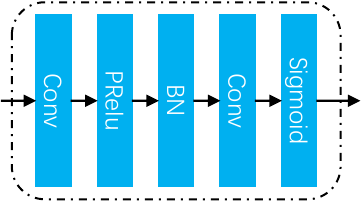} \label{fig:md}}\quad
    \subfloat[OPP]{\includegraphics[width=.37\linewidth]{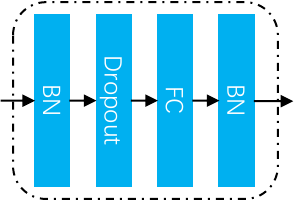} \label{fig:opp}}
    \caption{Architectures of MD and OPP. (a) MD represents the Mask Decoder. (b) OPP means the Occluded Pattern Predictor.}
    \label{fig:module}
\end{figure}

The principle of FROM is to learn a \textbf{Mask Decoder} to generate feature masks for precisely removing the corrupted features caused by occlusions. As illustrated in Figure~\ref{fig:framework}, it decodes occlusion information from conv feature maps $X_3$ obtained by the \textbf{Feature Pyramid Extractor} as the feature mask $M$. It is expected to mask out the corrupted feature elements of $X_1$ by an element-wise product to generate cleaned features $X^{'}_1$ for later recognition. As demonstrated by Figure~\ref{fig:md}, the \textbf{Mask Decoder} is implemented as a simple ``Conv-PReLU-BN-Conv-Sigmoid'' structure, in which the ``Sigmoid'' function is to constrain the output feature masks into $(0,1)$. Note that the stride of both ``Conv'' layers is set to 2 for down-scaling $4 \times$ resolution to exactly match the size of $X_1$. In the following, we justify several crucial choices for the \textbf{Mask Decoder}.

\subsubsection{Mask Source: middle \vs~deep}
Wan \etal~\cite{wan2017occlusion} attempted to utilize middle-stage feature maps to obtain the masks, but its reported accuracy is limited. In contrast, Song~\etal~\cite{song2019occlusion} adopted the top conv features to predict the masks as the deep convolutional features contain more semantic information with less noise. Our goal is to learn the feature mask which is highly sensitive to the occlusion location of the input image, while accurately cleaning the deep corrupted feature elements for recognition. Thus, we propose to utilize middle and deep features to capture both the local and global information to precisely learn the feature mask, which is achieved by the \textbf{Feature Pyramid Extractor}. Specifically, as illustrated in Figure~\ref{fig:framework}, we have:
\begin{align} 
X_2 &=  conv(upsample(conv(X_1)) + conv(C_2)),\\ 
X_3 &=  conv(upsample(conv(X_2)) + conv(C_3)),
\end{align}
where $conv(X)$ is the convolutional operation on feature map $X$ and $upsample(X)$ means up-sampling the resolution of $X$ by $2$ . As a result, $X_3$ contains all the different levels of information from $C_1, C_2, C_3$, which is fed into the \textbf{Mask Decoder} for predicting the corresponding feature masks.

\subsubsection{Mask Location: conv \vs~fc}
\label{sec:conv_fc}
Since the top features are usually used as the face representation for downstream tasks, it is straightforward to clean the corrupted elements of top features to yield better recognition performance. However, what kinds of format (conv \vs~fc) we are supposed to choose? Suppose there are two face images from different people with the same area (\eg, eyes) blocked, and then the learned masks should be the same regardless of different identities because they are required to depend solely on the occlusion locations. However, the top fc features have completely lost the spatial information while keeping highly-related identity information. Hence, as shown in Figure~\ref{fig:framework}, we propose to apply the learned feature mask on the top conv features, followed by the last FC layer to produce face representations. We also empirically compare the performances between cleaning conv and fc features in Sec.~\ref{sec:fc_2d} to reveal the superiority of masking conv features.

\subsubsection{Mask Dimension: 3D \vs~2D}
\label{sec:2d_3d}
In the input image space, all occlusion patterns are 2D ($H \times W$). Nevertheless, we propose to employ the 3D mask ($C \times H \times W$) on the top conv features. In fact, features in different channels $C$ usually have quite different functions, which is mostly caused by convolution operators. In particular, a feature is formed by the linear combination of preceding convoluted feature maps. The convolution operators (\eg, $3 \times 3$) expand 2D corruption in the preceding maps anisotropically due to the differences in kernel weights. As a result, a linear combination of such anisotropically convoluted maps is no longer channel independent. Therefore, we choose to learn the 3D mask of shape $C \times H \times W$ to capture such channel dependency of feature corruptions. We also empirically explore the difference between using 2D and 3D masks in Sec.~\ref{sec:fc_2d}.

\subsubsection{Mask Format: dynamic \vs~static}
The most recent state-of-the-art method~\cite{song2019occlusion} adopts \textit{static} masks for removing corrupted features. Face is divided into $K \times K$ grids, and $(i,j)$ indexes into these $K^2$ grids. It learns a dictionary in advance, where the key is the occluded location $(i,j)$ and the value is the corresponding binary mask. At the testing stage, \cite{song2019occlusion} first employs an external occlusion detector~\cite{long2015fully} to find out which of the $K \times K$ grids are occluded, and then the corresponding binary mask is fetched to clean the corrupted conv features.

In contrast, our FROM produces masks \textit{dynamically}, by decoding the occlusion information from the pyramid feature maps. The benefits of this mechanism are four-fold. First, at the testing phase, we only require one pass to extract occlusion-robust features because face feature extraction and mask generation share the same backbone network. But for static mask, it is required to detect occlusion with another deep neural network so that a mask corresponding to the testing occlusion can be retrieved. Second, the static method requires to train $K^2$ deep models to obtain a static mask dictionary, which is computationally expensive. Instead, our FROM only needs one single model training. Third, at testing dynamic masks adapt better to the particular testing instance which can have perturbations in occlusion locations or variations of the occluders. Finally, dynamic masks enable FROM to generalize well to clean face images.

\subsection{Occlusion Pattern Predictor}
\label{section:olp}

In order to encourage the \textbf{Mask Decoder} to generate masks related to the occlusion patterns of input face images, we introduce the module \textbf{Occlusion Pattern Predictor} into our network to supervise the feature masks learning. As illustrated by Figure~\ref{fig:framework}, it takes the learned feature mask as input and predicts the occlusion pattern vector which is then classified with softmax loss. It has a simple ``BN-Dropout-FC-BN'' structure with its output channel equal to the number of occlusion patterns as in Figure~\ref{fig:opp}. Therefore, the \textbf{Mask Decoder} is trained to generate masks: 1) related to occlusions of input images; and 2) correctly masking out corrupted features that are harmful to face recognition. While the first point is achieved by the introduction of the \textbf{Occlusion Pattern Predictor}, the second point is supervised by a face recognition loss (CosFace loss~\cite{wang2018cosface} in this paper) applied on the masked features.

\subsubsection{Proximate Occlusion Patterns}
\label{sec:proximity}
The face image is divided into $K \times K$ blocks and each block represents a sub-region of the face that may be occluded. Song~\etal~\cite{song2019occlusion} regarded each block as an independent one and merged multiple masks by logical AND if multiple blocks are occluded. The logical AND is an approximation to inferring complicated occlusion patterns. Without such an approximation, the number of different occlusions grows exponentially and becomes intractable quickly. In particular, if a face image is divided into $K \times K$ blocks, then there are $2^{K \times K}$ (each block may be occluded or not) different occlusion patterns.

\begin{figure}[t]
\centering
\includegraphics[width=0.75\linewidth]{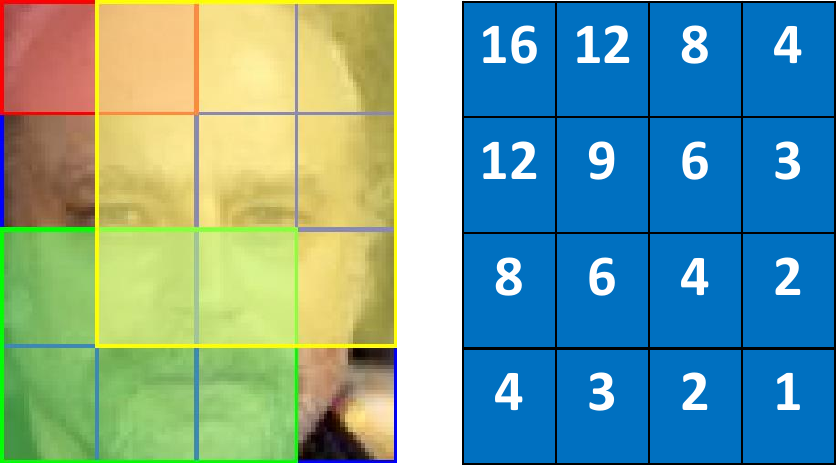}
\caption{Examples of proximate occlusion patterns (left) and the number of patterns for each occlusion size (right) when the face image is divided into $4 \times 4$ blocks. The value of $(i,j)$ location in the numerical matrix means the number of occlusion patterns with size $i \times j$.}
\label{fig:grid_division}
\end{figure}

In this section, we introduce another more intuitive approximation to overcome the above dimensionality issue. Observing that the adjacent blocks usually have similar occlusion statuses in practical applications (\eg, if the mouth is occluded then the nose is also occluded with high probability). We refer to this characteristic as proximity. Inspired by this characteristic, we reduce the number of possible occlusion patterns by restricting all occlusion patterns as rectangles to cover adjacent $m \times n$ blocks, where $m, n \in [1, K]$ represent the height and width of the occlusion pattern. Several occlusion patterns for the combination of $m$ and $n$ when $K = 4$ are displayed in Figure~\ref{fig:grid_division}. For example, the red pattern contains adjacent $1 \times 2$ blocks while the yellow pattern includes $3 \times 3$ blocks. The numerical matrix in Figure~\ref{fig:grid_division} indicates the numbers of different patterns when we divide the face images into $4 \times 4$ blocks. Specifically, the value of $(i, j)$ location in the matrix means the number of occlusion patterns with size $i \times j$ (\ie, covering adjacent $i \times j$ blocks). For instance, the value of $(2, 2)$ is $9$, indicating that there are 9 different patterns in total which cover adjacent $2 \times 2$ blocks. 

For the clean image, we regard its occlusion pattern as covering adjacent $0 \times 0$ blocks. By summarizing all the values of the matrix, we then have $101$ kinds of occlusion patterns ($100$ occluded and additional $1$ non-occluded). It is worth noting that the special occlusion pattern for the clean image makes FROM not rely on other networks like FCNs \cite{long2015fully} as in PDSN \cite{song2019occlusion} to detect whether the input image is occluded or not. In particular, the \textbf{Mask Decoder} will produce the harmless masks to the original features for clean images, making FROM generalize very well on general face recognition, which will be demonstrated by our experiments.

More generally, we have $(K \times (K+1)/2)^2 + 1$ kinds of occlusion patterns in total when dividing a face image into $K \times K$ blocks. We will experimentally discuss the impact of different choices of $K$ in Sec.~\ref{sec:pattern}.

\subsubsection{Pattern Prediction}
\begin{figure}[t]
\centering
\includegraphics[width=.95\linewidth]{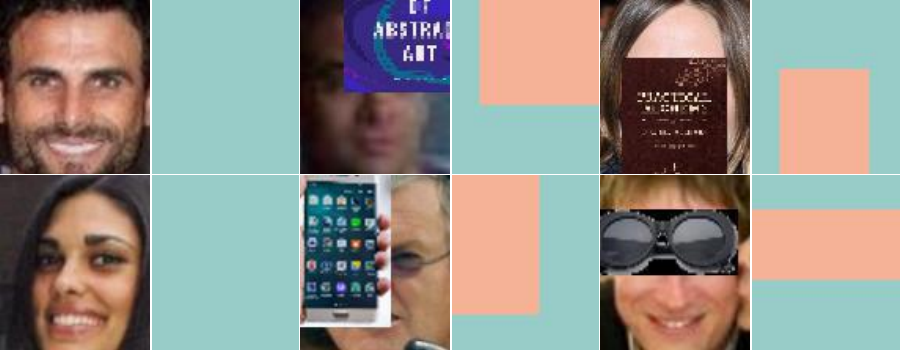}
\caption{Examples of input face images and their occlusions predicted by our FROM, where orange color indicates the occluded area. The first row is from Occ-WebFace, and the last row images are from Occ-LFW-1.5. By decoding the correct masks of the corresponding occlusion, corrupted features can be effectively removed.}
\label{fig:occ_label}
\end{figure}

The generated feature masks are supposed to be used to correctly predict the corresponding reference pattern. In particular, at the training phase, we have the occlusion location of each image. For each image $X_i$, we obtain its occlusion pattern $Y_i$ by matching its occlusion location to the $226$ (when $K = 5$) occlusion patterns. Our matching strategy is to calculate the IoU score between the occlusion and the $226$ reference patterns, and then pick the pattern with the largest IoU score as the corresponding label. After obtaining the occlusion feature vector from the \textbf{Occlusion Pattern Predictor}, we employ the traditional softmax loss with its occlusion pattern $Y_i$, which is formulated as follows:
\begin{equation}
 \mathcal{L}_{softmax} =  \frac{1}{N}\sum_{i=1}^N{-\log{p_i}} = \frac{1}{N}\sum_{i=1}^N{-\log{\frac{e^{f_{y_i}}}{\sum_{j=1}^C{e^{f_j}}}}},
 \label{eq:pred_loss}
\end{equation}
where $p_i$ denotes the posterior probability of $x_i$ being correctly classified. $N$ is the number of training samples and $C$ is the number of classes. $f$ is denoted as the occlusion feature vector. Figure~\ref{fig:occ_label} shows some examples of input face images and their predicted occlusion patterns.

\subsubsection{Pattern Regression}
\label{sec:p_reg}
Instead of formulating the pattern prediction as a classification problem, another straightforward idea is to directly regress the 2D coordinates of the occluded area without blocks division. Specifically, at the training stage, we have the rectangle location of occlusion for each face image, and directly take the 2D coordinates of upper-left and lower-right corners as the label $r$. Therefore, we require the vector $f$ obtained by the \textbf{Occlusion Pattern Predictor} be 4D to represent the coordinates. Then we employ the $L_2$ loss between $f$ and $r$ to guide the module learning, which can be formulated as follows:
\begin{equation}
 \mathcal{L}_{reg} =  \frac{1}{N}\sum_{i=1}^N{\| r - f\|_2},
 \label{eq:reg_loss}
\end{equation}
where $N$ is the number of training samples. Comparing to regarding pattern prediction as a classification task, directly regressing the coordinates of occlusion can avoid the quantization error brought by block division. However, the regression task is more difficult, which may shift the model training from the way of improving the recognition performance due to the sharing backbone network. We have empirically evaluated their differences in Sec.~\ref{sec:reg}.

\subsection{Overall Training Objective}
The overall loss is a combination of the face recognition loss and the occlusion pattern prediction loss, as shown in Figure~\ref{fig:framework}. Mathematically, we define the total loss as follows:
\begin{equation}
\label{eq:total_loss}
    \mathcal{L}_{total} = \mathcal{L}_{margin} + \lambda \mathcal{L}_{pred},
\end{equation}
where $\mathcal{L}_{margin}$ is defined by Eq.~\eqref{eq:margin} with $(m_1, m_2, m_3)$ being $(0, 0, m_3)$ (\ie, the CosFace loss \cite{wang2018cosface}), and $\mathcal{L}_{pred}$ is regarded as either Eq.~\eqref{eq:pred_loss} or Eq.~\eqref{eq:reg_loss}. Note that $\lambda$ is the weight coefficient of the occlusion pattern prediction loss, which will be explored in Sec.~\ref{sec:lamda}.

\begin{figure}[t]
	\centering
	\includegraphics[width=.9\linewidth]{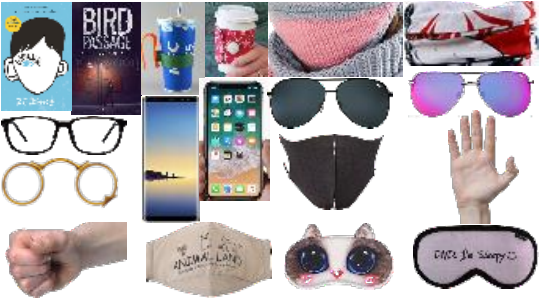}
	\caption{Illuminations of occluders we employ to construct the occluded face datasets. All of them are the most common objects that may block the face, including sunglasses, scarf, face mask, hand, eye mask, eyeglasses, book, phone, and cup.}
	\label{fig:occluder}
\end{figure}

\section{Experiments}
\subsection{Datasets and Evaluation Metrics}
\label{section:dataset}
\textbf{Datasets:} The face datasets we adopt for training and testing models are introduced as follows:

\vspace*{1.5mm}
\noindent
\textbf{\textit{CASIA-WebFace}~\cite{yi2014learning}:} The WebFace dataset includes various face images collected from the Internet and contains around 10,000 different subjects and in total nearly 500,000 images. Following~\cite{song2019occlusion}, the WebFace is used as the training dataset.

\vspace*{1.5mm}
\noindent
\textbf{\textit{Labeled faces in the wild (LFW)}~\cite{huang2008labeled}:} LFW is a standard face verification benchmark dataset which provides face images in unconstrained environments. We evaluate our model strictly following the standard protocol as in \cite{deng2019arcface} and report the verification performance on the 6,000 testing image pairs.

\vspace*{1.5mm}
\noindent
\textbf{\textit{Megaface Challenge 1 (MF1)}~\cite{kemelmacher2016megaface}:} MF1 is usually used for face identification task, where the gallery set contains more than one million face images as distractors, and the probe set includes two datasets: Facescrub~\cite{ng2014data} and FGNet. Following~\cite{song2019occlusion}, we adopt the Facescrub dataset as our probe set. Given that our model is trained on the WebFace dataset with $0.49M$ images, the protocol \textit{small} is used for evaluation.

\begin{figure}[t]
	\centering
	\includegraphics[width=.95\linewidth]{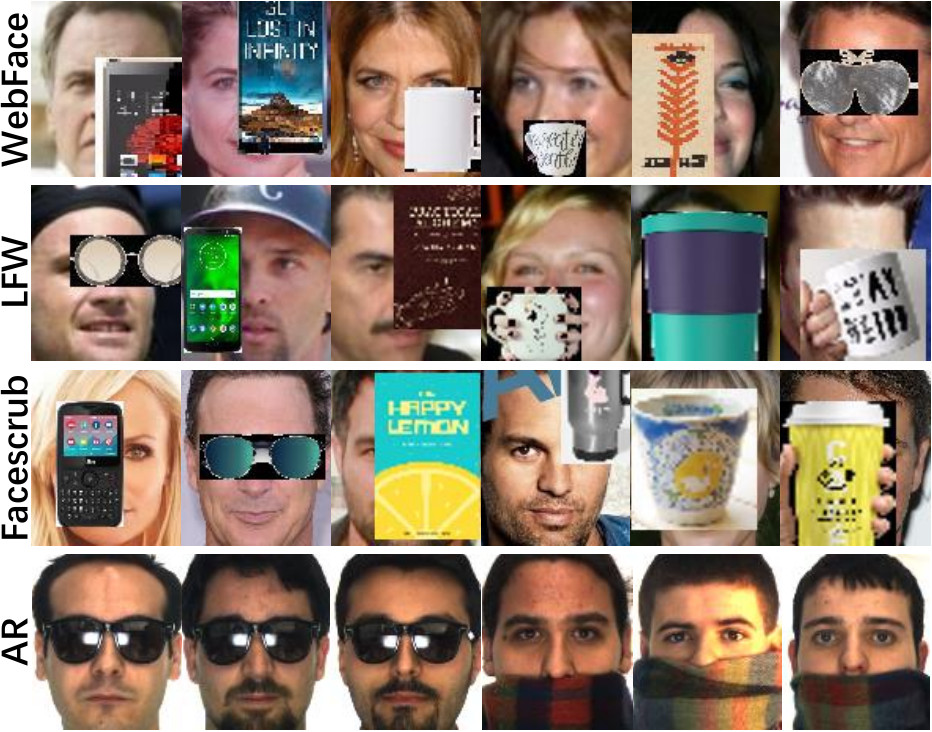}
	\caption{Examples of the AR Face dataset and our synthetically occluded face images. The first row images are occluded with random scale from $1:0.5:5$ on WebFace. The second and third rows images are occluded with $1.0, 1.5, 2.0$ on LFW and Facescrub, respectively. The last row images are from the AR Face dataset.}
	\label{fig:occ_imgs}
\end{figure}

\vspace*{1.5mm}
\noindent
\textbf{\textit{AR Face Dataset}~\cite{martinez1998ar}:} The AR face dataset contains over 4,000 face images with different facial expressions, illumination conditions and occlusions from 126 persons. It includes two kinds of occluded face images sets. People wear sunglasses or scarf when they were pictured to form ``Sunglasses'' and ``Scarf'' probe sets, respectively. Face identification evaluation is performed on AR dataset under two different protocols. Some examples are visualized in Figure~\ref{fig:occ_imgs}.

\vspace*{1.5mm}
\noindent
\textbf{\textit{Synthetic Datasets}:} Following \cite{song2019occlusion}, we synthesize occluded datasets using the aforementioned datasets with occluders including sunglasses, scarf, face mask, hand, eye mask, eyeglasses, book, phone, and cup, all of which are common objects in real-life that may appear on the face. Several examples are illustrated in Figure~\ref{fig:occluder}. The occluded face images are formed by pasting random occluders on the random location of faces, which is detailedly demonstrated by Alg.~\ref{alg:construct}. It is worth noting that some images might be occluded by the part of occluder because we only require the center of occluder inside the image area instead of the whole occluder. We name the generated occluded face datasets as ``Occ-D-S'', where ``D'' means the original face dataset (\eg, LFW, Facescrub from MF1) and ``S'' is Scale $s$ of Alg.~\ref{alg:construct}  (\eg, $1.0, 2.0$). Note that $s$ can also be regarded as the occluded degree, \eg, $s=1.0, 2.0, 3.0$ correspond to the occluded area are $18.76\%, 43.09\%, 52.00\%$ of the total face image area respectively when applied on LFW with the random seed setting to 100. Some synthesized face images with occlusions are illustrated in Figure~\ref{fig:occ_imgs}.

\begin{algorithm}[!htb]
\begin{algorithmic}
\STATE \textbf{Input:} Face Image $img$, Random Occluder $occ$, Scale $s$.
\vspace{-1mm}
\begin{enumerate} \setlength{\itemsep}{-\itemsep}
\item obtaining width and height $(occ\_w, occ\_h)$ of $occ$
\item rescaling $occ$ to $(s*occ\_w, s*occ\_h)$ size 
\item picking a random point of $img$ as the center of $occ$
\item calculating the occluded area $box$ based on the center and size of $occ$ 
\item pasting $occ$ on the occluded $box$ to obtain $occ\_img$ 
\end{enumerate}
\vspace{-1mm}
\STATE \textbf{Output:} Occluded Face Image $occ\_img$ and Area $box$ 
\end{algorithmic}
\caption{Occluded Face Datasets Construction}
\label{alg:construct}
\end{algorithm}

\vspace*{2mm}
\noindent
\textbf{Evaluation Metrics:} We employ two widely-used metrics to quantitatively evaluate the face recognition performance in our experiments. The first one is the accuracy metric:
\begin{equation}
  Accuracy =  \frac{TP + TN}{TP + TN + FP + FN},
 \label{eq:acc}
\end{equation}
where $TP$, $TN$, $FP$ and $FN$ are true positives, true negatives, false positives and false negatives, respectively. The second metric we adopt is TAR (True Accepted Rate) under FAR (False Accepted Rate), which can be defined as:

\begin{equation}
    \begin{split}
        TAR = \frac{TP}{TP + FN}, \quad FAR = \frac{FP}{FP + TN}.
    \end{split}
    \label{eq:tar}
\end{equation}

\subsection{Implementation Details}
\label{section:implement}
\textbf{Pre-processing.} 
Firstly, we detect five landmarks (two eyes, nose and two mouth corners) for each face image with MTCNN \cite{zhang2016joint}. Then we perform corresponding similarity transformation to align and crop face images properly to obtain face images with $112 \times 96$ size. Example images are shown in Figure~\ref{fig:occ_imgs}. Following \cite{song2019occlusion, wang2018cosface}, the pixel values are normalized to $[-1.0, 1.0]$ when training and testing.

\vspace*{2mm}
\noindent
\textbf{Training.}
The training phase can be divided into two steps. First, we learn the backbone network (\ie, the upper branch in Figure~\ref{fig:framework}) on the WebFace dataset with the large margin cosine loss \cite{wang2018cosface} for general face recognition. The backbone network is trained for 40 epochs with the initial learning rate 0.1 and batch size 512. SGD with weight decay $0.0005$ and momentum $0.9$ is used, and the learning rate is reduced by $10$ times at epoch 15 and 30. Then we take the trained model from step 1 as our pretrained model and finetune the whole network including the \textbf{Feature Pyramid Extractor}, \textbf{Mask Decoder}, and \textbf{Occlusion Pattern Predictor} on the Occ-WebFace dataset, which is constructed by randomly choosing from $1.0:0.5:5.0$ as Scale $s$ in Alg.~\ref{alg:construct} and occluders for each image from the original WebFace dataset.

\vspace*{2mm}
\noindent
\textbf{Baseline Settings.}
In this paper, three important baselines are considered: 

\vspace*{1.5mm}
\noindent
Baseline: the backbone network (\ie, the upper branch in Figure~\ref{fig:framework}) only trained on the WebFace dataset (clean face images) with the large margin cosine loss~\cite{wang2018cosface}.

\vspace*{1.5mm}
\noindent
Baseline-Aug: the second baseline is derived from the first baseline by occluded data augmentation. It takes the trained Baseline as pretrained model and finetunes on Occ-WebFace for 25 epochs with the initial learning rate 0.01. SGD with weight decay $0.0005$ and momentum $0.9$ is used, and the learning rate is reduced by $10$ times at epoch 15.

\vspace*{1.5mm}
\noindent
Baseline-MD: the third baseline, equipping with the \textbf{Mask Decoder} and taking the trained Baseline as pretrained model, is trained on the Mix-WebFace dataset which is the mixup of Occ-WebFace and WebFace with the ratio $2:1$ for 40 epochs. The initial learning rate is $0.01$ and batch size is 512. SGD with weight decay $0.0005$ and momentum $0.9$ is used, and the learning rate is divided by $10$ at epoch 10 and 25. The difference between Baseline-MD and FROM is that it does not have the \textbf{Occlusion Pattern Predictor} to guide the \textbf{Mask Decoder} to learn the proper masks, \ie, $\lambda = 0$ in Eq.\eqref{eq:total_loss}. Its supervision is solely from the face recognition loss $\mathcal{L}_{margin}$.

\subsection{Ablation Study}

\subsubsection{Mask Binarization}
The feature masks produced by the \textbf{Mask Decoder} have continuous values within $(0, 1)$ due to the ``Sigmoid'' function, which can be regarded as soft masks when compared to the hard binary masks (\ie, only contain $0, 1$). To explore their difference, we apply binarization with a threshold $t \in (0,1)$ to convert the learned continuous mask values $x$ into binary values $B(x)$: if $x \geq t$, $B(x)$ will be $1$, otherwise it will be $0$. We compare their face verification accuracies on LFW with different thresholds as $t = 0.3:0.05:0.6$. As shown in Table~\ref{tab:binarization}, we can observe that the soft version of feature masks achieves the best accuracy on both LFW and Occ-LFW-2.0 with $99.38\%$ and $94.70\%$, and the hard version obtains similar performances only when threshold $t \leq 0.5$. It suggests that suppressing the corrupted feature elements is better than totally removing them, because they may still contain minor information for recognition. Another phenomenon deserved to discuss is that the performances by hard binary masks on LFW are nearly unchanged when threshold $t \leq 0.5$, which reveals the most elements of learned masks for clean images are high values that ensures our FROM generalizes well on clean face image recognition. From the perspective of performance and convenience, we adopt the learned soft feature masks for the rest of experiments.

\begin{table*}[t]
\renewcommand\arraystretch{1.5}
\centering
\caption{Face verification accuracy (\%) comparisons between no binarization and binarization with different thresholds under $K=5, \lambda=1.0$.}
\label{tab:binarization}
\vspace{-1mm}
\begin{tabular}{|p{2.5cm}<{\centering} | p{2.5cm}<{\centering} | p{1cm}<{\centering}| p{1cm}<{\centering} | p{1cm}<{\centering} | p{1cm}<{\centering}| p{1cm}<{\centering} | p{1cm}<{\centering}| p{1cm}<{\centering}|}
\hline
\multirow{2}*{Datasets} & \multirow{2}*{No-Binarization} & \multicolumn{7}{c|}{Binarization} \\
\cline{3-9}
& & 0.30 & 0.35 & 0.40 & 0.45 & 0.50 & 0.55 & 0.60 \\ \hline
LFW & \textbf{99.38} & 99.38 & 99.33 & 99.42 & 99.30 & 99.30 & 99.07 & 96.62 \\ \hline
Occ-LFW-2.0 & \textbf{94.70} & 94.17 & 94.40 & 94.35 & 94.40 & 93.83 & 92.78 & 87.32 \\ \hline
\end{tabular}
\end{table*}

\subsubsection{Discussion of $K$}
\label{sec:pattern}
When a face image is divided into $K \times K$ grids, the number of occlusion patterns increases exponentially (\ie, $2^{K^2}$). For example, When $K =4$, if we directly adopt $2^{4 \times 4}$ occlusion patterns in the model, then we have $65536$ patterns with $1456.8M$ parameters in total, which makes it hard and time-consuming to train. In contrast, the parameters are significantly reduced to $101$ patterns with $49.55M$ parameters by employing our approximation as described in Sec.~\ref{sec:proximity}. 

\begin{table}[t]
\center
\vspace{-3mm}
\caption{Face identification accuracies(\%) for different values of $K$ on MF1 and ``Occ-MF1-S'' with S $\in [1.0, 1.5, 2.0]$. ``Avg'' averages on three different ``Occ-MF1-S'' excluding MF1.}
\vspace{-2mm}
\label{table:kvalue_mf}
\begin{tabular}{p{0.4cm}  p{0.5cm}<{\centering} p{1.58cm}<{\centering} p{1.58cm}<{\centering} p{1.58cm}<{\centering} p{0.5cm}<{\centering}}
\hline
\noalign{\smallskip}
$K$  & MF1  & Occ-MF1-1.0  & Occ-MF1-1.5  & Occ-MF1-2.0 & Avg   \\
\noalign{\smallskip}
\hline
\noalign{\smallskip}
$K_0$ & 73.64 &	32.00 &	14.62 &	7.70 &	18.11   \\
\noalign{\smallskip}
$K_3$ & 74.04 & 59.64 & 41.72 & 30.07 & 43.81  \\
\noalign{\smallskip}
$K_4$ & 74.22 &	60.31 &	42.30 &	30.49 &	44.37  \\
\noalign{\smallskip}
$K_5$  & \textbf{75.27} &	\textbf{60.84} &	42.67 &	30.96 &	44.83 \\
\noalign{\smallskip}
$K_6$  & 74.65 &	60.62 &	\textbf{42.96} &	\textbf{31.15} &	\textbf{44.91}  \\
\hline
\end{tabular}
\vspace{-2mm}
\end{table}

We explore how different choices of $K$ will impact the final recognition accuracy by setting $K \in [3, 4, 5, 6]$ with fixing the weight coefficient $\lambda=1$ from Eq.\eqref{eq:total_loss}. We test all of the models on MF1 and ``Occ-MF1-S'' with S $ \in [1.0, 1.5, 2.0]$, respectively. As shown in Table~\ref{table:kvalue_mf}, $K_5$ obtains the highest accuracy $75.27\%$ on MF1, which is state-of-the-art performance under protocol \textit{small} and even better than $K_0$ (\ie, the \textit{Baseline} described in Sec.~\ref{section:implement}).  When we replace the original clean probe set with occluded ones, the accuracy is significantly degraded. $K_5$ degrades from  $75.27\%$ to $60.84\%$ on Occ-MF1-1.0, which, however, is still tremendously surpasses the $K_0$($32.00\%$). It sufficiently proves the effectiveness of FROM. Finally, $K_4, K_5, K_6$ achieves averagely similar performances while $K_3$ slightly falls behind. Considering speed and accuracy tradeoff, we adopt $K=5$ (\ie, dividing face image into $5 \times 5$ grids) for the remaining experiments.

\begin{table}[!htb]
\center
\caption{Face identification accuracies(\%) for different values of $\lambda$ (\ie, the weight coefficient of Eq.~\eqref{eq:total_loss}) on MF1 and ``Occ-MF1-S'' with S $\in [1.0, 1.5, 2.0]$. ``Avg'' averages on three different ``Occ-MF1-S'' excluding MF1.}
\vspace{-2mm}
\label{table:value}
\begin{tabular}{p{0.4cm}  p{0.5cm}<{\centering} p{1.58cm}<{\centering} p{1.58cm}<{\centering} p{1.58cm}<{\centering} p{0.5cm}<{\centering}}
\hline
\noalign{\smallskip}
$\lambda$ &MF1  & Occ-MF1-1.0  & Occ-MF1-1.5  & Occ-MF1-2.0 & Avg   \\
\noalign{\smallskip}
\hline
\noalign{\smallskip}
$\lambda_0$ & 74.17 &	59.31 &	41.38 &	29.74 &	43.48  \\
\noalign{\smallskip}
$\lambda_{0.5}$ & 74.88 &	60.61 &	\textbf{43.14} &	\textbf{31.26} &	\textbf{45.00}   \\
\noalign{\smallskip}
$\lambda_{1.0}$  & \textbf{75.27} &	\textbf{60.84} &	42.67 &	30.96 &	44.83 \\
\hline
\end{tabular}
\end{table}

\begin{table*}[t]
\center
\caption{Comparison between three baselines (Note: BN=Backbone Network, AUG=Occluded Data Augmentation, MD=Mask Decoder, OPP=Occlusion Pattern Predictor) and FROM. Note that the ``Occ-MF1-AVG'' means the average accuracy on the nine above occluded Facescrub. Face identification accuracies(\%) are reported on MF1 and Occ-MF1-AVG, while Face verification accuracies(\%) are reported on LFW and Occ-LFW-2.0.}
\vspace{-2mm}
\label{table:module}
\begin{tabular}{p{2cm}  p{1cm}<{\centering} p{1cm}<{\centering} p{1cm}<{\centering} p{1cm}<{\centering} p{1cm}<{\centering} p{2.5cm}<{\centering} p{1cm}<{\centering} p{2.5cm}<{\centering}}
\hline
\noalign{\smallskip}
Method  & BN  \quad & AUG  \quad & MD \quad & OPP  \quad & MF1 \quad & Occ-MF1-AVG  \quad & LFW \quad & Occ-LFW-2.0 \quad \\
\noalign{\smallskip}
\hline
\noalign{\smallskip}
Baseline & $\checkmark$ &  &  &  & 73.64 & 60.02  & 99.38 & 82.05\\
\noalign{\smallskip}
Baseline-Aug & $\checkmark$& $\checkmark$ &  &  & 72.50 & 67.91 & 99.10 & 94.10\\
\noalign{\smallskip}
Baseline-MD & $\checkmark$& $\checkmark$ & $\checkmark$ &  & 74.17 & 68.19  & 99.33 & 94.42\\
\noalign{\smallskip}
FROM  & $\checkmark$& $\checkmark$ & $\checkmark$ & $\checkmark$ & \textbf{75.27} & \textbf{69.85} & \textbf{99.38} & \textbf{94.70}\\
\hline
\end{tabular}
\vspace{-2mm}
\end{table*}

\subsubsection{Discussion of $\lambda$}
\label{sec:lamda}
The weight coefficient $\lambda$ of Eq.~\eqref{eq:total_loss} controls the tradeoff of face recognition loss and pattern prediction loss. Intuitively, if $\lambda$ is too small, the \textbf{Mask Decoder} lacks of effective supervision to learn accurate feature masks; if $\lambda$ is too large, then the model might focus on improving the accuracy of pattern prediction instead of recognition itself. Thus, we have explored the impact of different $\lambda$ based on the scale of face recognition and pattern prediction loss. We test $\lambda \in [0.0, 0.5, 1.0]$ under $K=5$ on MF1 and ``Occ-MF1-S'' with S $ \in [1.0, 1.5, 2.0]$, respectively, and the results are demonstrated in Table~\ref{table:value}. Note that $\lambda_0$ is corresponding to the \textit{Baseline-MD} described in Sec.~\ref{section:implement}. Apparently, $\lambda_{0.5}, \lambda_1$ obtain the excellent accuracies that surpass $\lambda_0$ by a clear margin, which suggests the effectiveness of the proposed \textbf{Mask Decoder} and pattern approximation.

\subsection{Comparison to Baselines}
In this experiment, we compare FROM to the three aforementioned baselines (described in Sec.~\ref{section:implement}) under the same configurations. Two different performances on occluded face datasets are reported. ``Occ-MF1-AVG'' indicates the average accuracy on the nine occluded types of Facescrub (as demonstrated by Figure~\ref{fig:diff}) from the Megaface Challenge 1. ``Occ-LFW-2.0'' means the occluded version of LFW by setting the Scale factor $S$ to $2$ in the construction algorithm~\ref{alg:construct}. In order to evaluate how well different methods are generalized to general face recognition without occlusions, we also compare the methods on the original LFW and MF1 datasets. 

The results shown in Table~\ref{table:module} reveal that occluded data augmentation is quite effective to improve accuracy for occluded face recognition. The Baseline-Aug has significantly improved the accuracy from the Baseline trained for general face recognition on both Occ-MF1-AVG and Occ-LFW-2.0, which is consistent with our intuition. Next, the Baseline-MD with the \textbf{Mask Decoder} performs sightly better than Baseline-Aug, which might be brought by the extra network parameters. However, when comparing our FROM against the Baseline-MD, we can see that it consistently outperforms Baseline-MD with a clear margin on all datasets, which reveals that the \textbf{Occlusion Pattern Predictor} guides the \textbf{Mask Decoder} well to learn proper masks to clean those corrupted feature elements. As for the generalization to general face datasets, we first find that Baseline-Aug performs slightly worse than Baseline, which might be caused by the overfitting problem in its occluded data finetune process. In contrast, FROM improves the accuracy from $73.64\%$ to $75.27\%$ on MF1 while keeping the high accuracy $99.38\%$ not degraded on LFW. It contributes to our special pattern design for clean face image inside the \textbf{Occlusion Pattern Prediction}. All above comparisons confirm that our FROM is effective for occluded face recognition, and generalizes well or even better on general face recognition.

\begin{figure}[t]
	\centering
	\includegraphics[width=.95\linewidth]{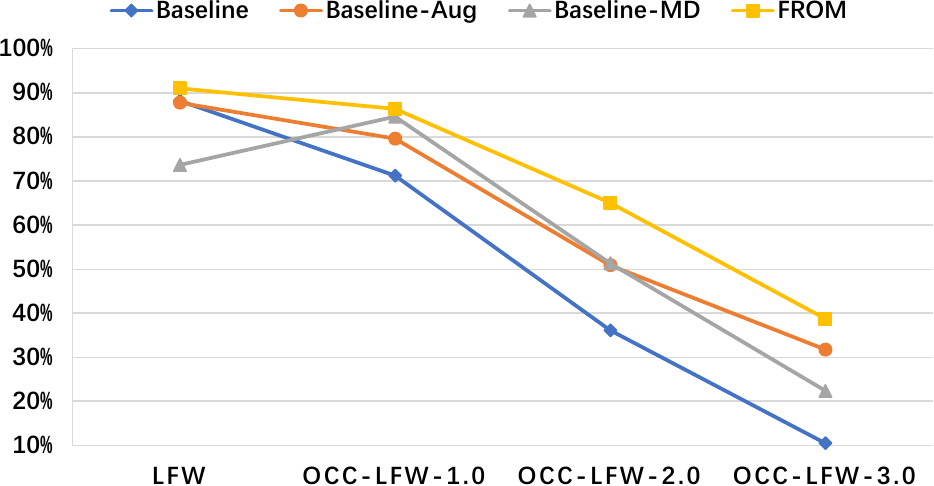}
	\caption{The values of TAR (True Accepted Rate) when FAR (False Accepted Rate) is $1e-4$ on LFW and  ``Occ-LFW-S'' with S $\in [1.0, 2.0, 3.0]$.}
	\label{fig:tar_far}
\end{figure}

The improvements on LFW and Occ-LFW-2.0 are limited from Table~\ref{table:module}, which we think are caused by the loose metric (\ie, verification accuracy). Hence, we propose to introduce a more strict metric to reveal the superiority of our FROM. We consider the TAR (True Accepted Rate) under specific FAR (False Accepted Rate) defined by Eq.~\eqref{eq:tar} on both LFW and ``Occ-LFW-S'' with S $\in [1.0, 2.0, 3.0]$. The FAR is set to $1e-4$ which is a very hard threshold, and then we obtain the TAR for three baselines and FROM as illustrated in Figure~\ref{fig:tar_far}. The Baseline-MD obtains the worst result on clean dataset LFW, and we think the potential reason is that the learned feature masks for clean face image by \textbf{Mask Decoder} are inaccurate due to the absence of supervision from \textbf{Occlusion Pattern Predictor}. Not surprisingly, FROM consistently outperform three baselines on both clean and occluded face datasets. Importantly, when testing dataset becomes harder, our FROM remarkably surpasses three baselines by a large margin (\eg, $65.03\%$ \vs~$50.87\%$ against Baseline-Aug on Occ-LFW-2.0).

\begin{figure}[t]
    \centering
    \vspace{-4mm}
    \subfloat{\includegraphics[width=.45\linewidth]{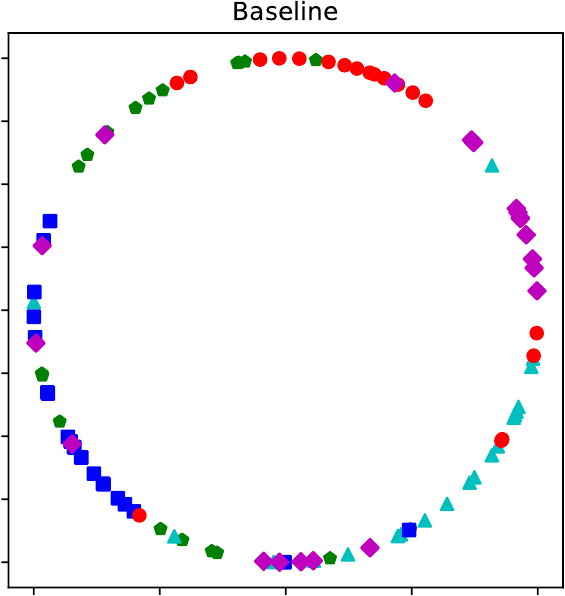} \label{fig:fd1}}\quad
    \subfloat{\includegraphics[width=.45\linewidth]{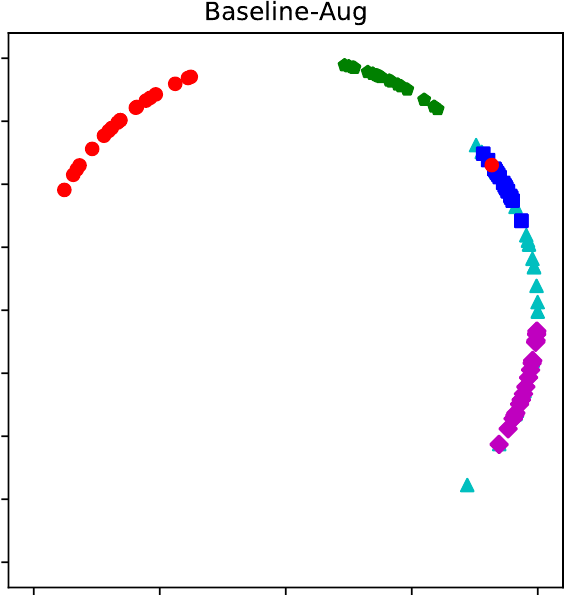} \label{fig:fd2}}\quad
    \subfloat{\includegraphics[width=.45\linewidth]{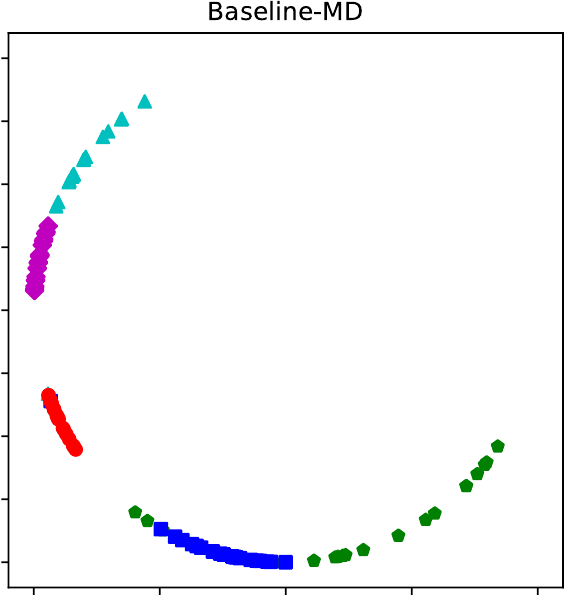} \label{fig:fd3}}\quad
    \subfloat{\includegraphics[width=.45\linewidth]{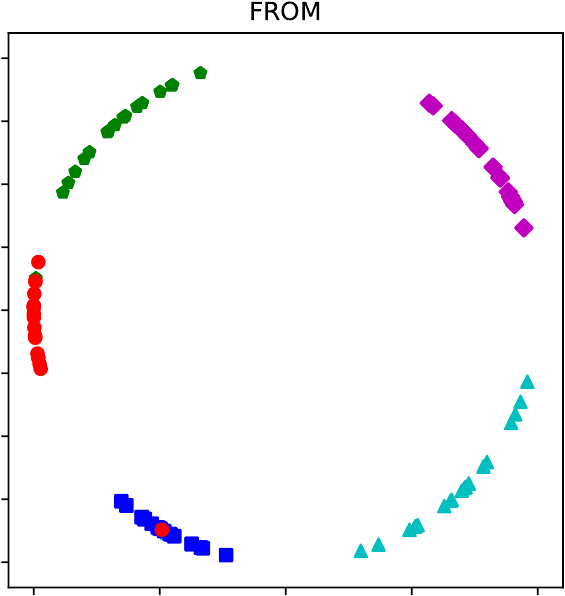} \label{fig:fd4}}
    \caption{Visualization of feature distributions by converting 512D to 2D with $t$-SNE~\cite{van2008visualizing} and following normalization. Different markers with color represent different classes. Zoom in for better view.}
    \label{fig:fd}
    \vspace{-2mm}
\end{figure}

To explore the improvement brought by FROM in depth, we compare the feature distributions used for recognition between aforementioned four models. 5 classes with 20 samples per class in Occ-LFW-2.0 are randomly picked, and then we obtain their 512D features for each model. After that, the features are converted to 2D by $t$-SNE~\cite{van2008visualizing} and visualized after normalization in Figure~\ref{fig:fd}. Obviously, Baseline cannot handle the severe occluded face images from Occ-LFW-2.0 as its distributions are in a mess. Baseline-Aug and Baseline-MD barely tell these classes apart with some classes still interlaced. However, FROM can distinguish these classes with clear margins, revealing its superiority.

\begin{table}[b]
\center
\vspace{-4mm}
\caption{Face identification accuracies(\%) on ``Occ-MF1'' with nine different occluded areas of face images. For example, ``upper'' refers to face images that are occluded by random occluders around the upper part of face. ``avg'' refers to the mean accuracy for all occluded areas.}
\vspace{-2mm}
\label{table:mf_sp}
\begin{tabular}{p{1.8cm}  p{1.2cm}<{\centering} p{0.7cm}<{\centering} p{0.7cm}<{\centering} p{0.7cm}<{\centering} p{1.2cm}<{\centering}}
\hline
\noalign{\smallskip}
Method  & left &	right &	upper &	lower &	left eye   \\
\noalign{\smallskip}
\hline
\noalign{\smallskip}
Baseline & 61.66 &	62.54 &	49.50 &	57.60 &	63.85   \\
\noalign{\smallskip}
Baseline-Aug & 67.30 &	68.94 &	\textbf{61.72} &	68.44 &	70.20    \\
\noalign{\smallskip}
Baseline-MD  & 67.66 &	69.90 &	59.02 &	\textbf{69.86} &	70.62  \\
\noalign{\smallskip}
FROM  & \textbf{68.98} &	\textbf{70.60} &	61.37 &	69.72 &	\textbf{73.51}  \\
\noalign{\smallskip}
\hline
\noalign{\smallskip}
Method &	right eye &	eyes &	nose &	mouth &	avg \\
\noalign{\smallskip}
\hline
\noalign{\smallskip}
Baseline & 	66.63 &	51.39 &	56.67 &	70.31 &	60.02   \\
\noalign{\smallskip}
Baseline-Aug & 	70.96 &	64.74 &	67.88 &	71.04 &	67.91   \\
\noalign{\smallskip}
Baseline-MD  & 	72.20 &	63.30 &	68.55 &	72.58 &	68.19  \\
\noalign{\smallskip}
FROM  &	\textbf{74.09} &	\textbf{65.40} &	\textbf{70.99} &	\textbf{73.99} &	\textbf{69.85}  \\
\hline
\end{tabular}
\vspace{-2mm}
\end{table}

\subsection{Effects of Different Occluded Areas}
\label{sec:affect}
Intuitively, occlusions on crucial areas (\eg, eyes) affect the recognition accuracy more significantly. In this experiment, we quantitatively evaluate how different occluded areas impact the accuracy based on the MF1 dataset. In particular, we explore nine types of commonly encountered occlusion types, including half face, eyes, nose, mouth, which are illustrated in Figure~\ref{fig:diff}. The results of the three baseline methods are included to make the study more informative, as shown in Table~\ref{table:mf_sp}. It is interesting that occluded areas affect different methods quite consistently. For example, occlusion on the mouth does not degrade too much accuracy, while occlusion on eyes reduces the accuracy quite significantly. This confirms that certain facing regions (\eg, eyes) contain more identity information than other regions (\eg, nose and mouth). Another observation is that models perform quite similarly on the left and right face, which might contribute to the flip augmentation for training face images. Last but not least, our FROM significantly improves the accuracy over \textit{Baseline} under all the occlusion types again, which shows the excellence of our method.

\subsection{Effects of Different Mask Locations and Dimension}
\label{sec:fc_2d}

\begin{table}[b]
\center
\vspace{-3mm}
\caption{Face identification accuracies(\%) of different masks locations and dimensions on MF1 and ``Occ-MF1-S'' with S $\in [1.0, 1.5, 2.0]$. Note that ``FC'' represents applying the masks on fc features; ``2D'' and ``3D'' mean the mask dimensions are 2D and 3D. ``3D'' is corresponding to our FROM. ``Avg'' averages on three different ``Occ-MF1-S'' excluding MF1.}
\label{table:fc_2d}
\vspace{-2mm}
\begin{tabular}{p{0.4cm}  p{0.5cm}<{\centering} p{1.58cm}<{\centering} p{1.58cm}<{\centering} p{1.58cm}<{\centering} p{0.5cm}<{\centering}}
\hline
\noalign{\smallskip}
F & MF1  & Occ-MF1-1.0  & Occ-MF1-1.5  & Occ-MF1-2.0 & Avg   \\
\noalign{\smallskip}
\hline
\noalign{\smallskip}
FC & 74.97 &	59.75 &	42.13 &	30.26 &	44.05   \\
\noalign{\smallskip}
2D & 74.52 &	60.56 &	42.36 &	30.13 &	44.35   \\
\noalign{\smallskip}
3D  & \textbf{75.27} &	\textbf{60.84} &	\textbf{42.67} &	\textbf{30.96} &	\textbf{44.83} \\
\hline
\end{tabular}
\vspace{-2mm}
\end{table}

\begin{figure}[b]
	\centering
	\includegraphics[width=.95\linewidth]{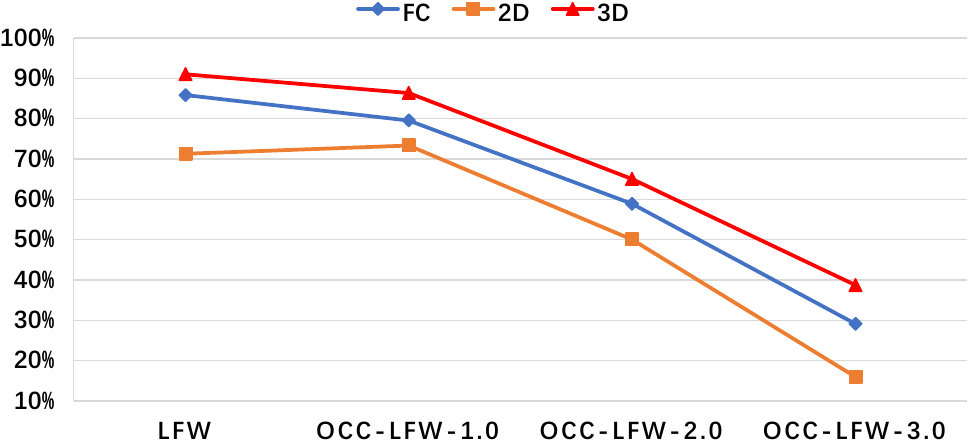}
	\caption{The values of TAR (True Accepted Rate) when FAR (False Accepted Rate) is $1e-4$ on LFW and  ``Occ-LFW-S'' with S $\in [1.0, 2.0, 3.0]$ from FC, 2D and 3D. FROM is corresponding to ``3D''.}
	\label{fig:fc_2d}
\end{figure}

In this experiment, we first explore the effectiveness of masking out the fc features instead of conv features. As we discuss in Sec.~\ref{sec:conv_fc}, the learned masks should solely depend on the location of occlusions instead of its types and the identity of input face image. However, the top fc features have totally lost all the spatial information while keeping high-related identity information. Thus, masking out the fc features may degrade the performance. We first evaluate the models on MF1 and ``Occ-MF1-S'' with S $\in [1.0, 1.5, 2.0]$ as shown in Table~\ref{table:fc_2d}. Our FROM (\ie, ``3D'') consistently outperforms the ``FC'', but the improvement is limited. Thus, we introduce TAR under FAR again as demonstrated in Figure~\ref{fig:fc_2d}. Under harder metric, our FROM (\ie, ``3D'') remarkably surpasses the ``FC'', confirming our above conjecture.

We then discuss the impacts of mask dimensions (\ie, 2D \vs~3D). As we mention in Sec.~\ref{sec:2d_3d}, the deep conv feature maps are no longer channel independent. If we apply same 2D masks across all of the channels, the performance might be decreased. Similarly, we carry experiments on ``Occ-MF1-S'' with face identification accuracy, LFW and ``Occ-LFw-S'' with TAR under specific FAR. As demonstrated by Figure~\ref{fig:fc_2d},  Our FROM (\ie, ``3D'') consistently outperforms the ``2D'' by a large margin, which suggests the 3D masks can capture the channel dependency of features to obtain more accurate feature masks, yielding better recognition accuracy.

\subsection{Pattern Regression \vs~Classification}
\label{sec:reg}
Intuitively, directly regressing the occluded area can avoid the quantization error caused by block division to obtain more accurate masks. However, comparing to the pattern classification we propose, direct pattern regression is more difficult which may degrade the accuracy of recognition. In this experiment, we explore the effectiveness of pattern regression as defined in Sec.~\ref{sec:p_reg}. We adopt the TAR when FAR is $1e-3$ to evaluate the performances between ``REG'' and FROM, which is demonstrated in Table~\ref{table:reg}. Note that the footnote on ``REG'' indicates the value of weight coefficient $\lambda$ (\eg, ``$REG_{0.1}$'' means $\lambda=0.1$ for training). Apparently, $\lambda=0.1$ achieves the best accuracy among ``$REG_{\lambda}$'', but FROM surpasses it on all datasets. Specifically, when testing dataset becomes harder, the gap is larger, which confirms the effectiveness of the proposed pattern approximation.

\begin{table}[t]
\center
\vspace{-2mm}
\caption{Comparison of TAR (True Accepted Rate) when FAR (False Accepted Rate) is set to $1e-3$ between the regression loss and our FROM.}
\vspace{-2mm}
\label{table:reg}
\begin{tabular}{p{1cm}  p{0.5cm}<{\centering} p{1.7cm}<{\centering} p{1.7cm}<{\centering} p{1.7cm}<{\centering} }
\hline
\noalign{\smallskip}
Method & LFW  & Occ-LFW-1.0  & Occ-LFW-2.0  & Occ-LFW-3.0  \\
\noalign{\smallskip}
\hline
\noalign{\smallskip}
FROM & \textbf{98.63} & \textbf{96.17} & \textbf{76.53} &  \textbf{70.23}  \\
\noalign{\smallskip}
$REG_{0.1}$ & 98.33 & 94.6 & 74.93 &  66.57   \\
\noalign{\smallskip}
$REG_{0.2}$ & 79.33 & 70.33 & 47.40 &  30.50   \\
\noalign{\smallskip}
$REG_{0.5}$ & 45.10 & 41.87 & 22.77 &  17.03   \\

\hline
\end{tabular}
\vspace{-2mm}
\end{table}

\begin{figure}[t]
	\centering
	\includegraphics[width=.9\linewidth]{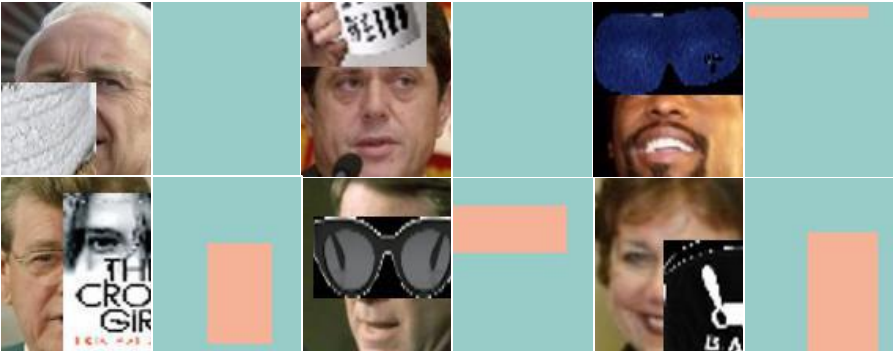}
	\vspace{-2mm}
	\caption{Visualizations of predicted masks by directly regressing. As we can observe, some of the regressed masks are not accurate, and the first two images are even miss-detected.}
	\label{fig:reg}
	\vspace{-3mm}
\end{figure}

Figure~\ref{fig:reg} illustrates several regression results by ``$REG_{0.1}$''. Some regressed masks are not accurate, and the first two images are even miss-detected. We think the reason is that the weight coefficient $\lambda=0.1$ is too small, leading to less attention on pattern regression. However, when we increase the $\lambda$ to $0.2, 0.5$, the face verification accuracy is significantly degraded as illustrated in Table~\ref{table:reg}. Therefore, our FROM with pattern approximation is the superiority 

\subsection{Generalization on Unseen Occlusions}
\begin{figure}[t]
	\centering
	\includegraphics[width=.85\linewidth]{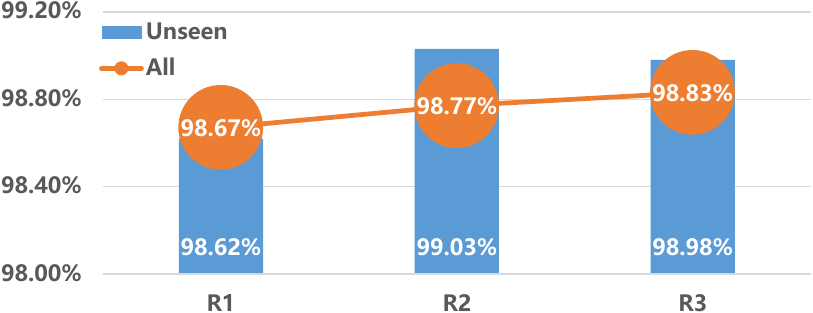}
	\vspace{-2mm}
	\caption{Generalized accuracies on unseen occlusions under $S=1, K=4, \lambda=1.0$. ``R1'' is trained with the dataset occluded by the first group and evaluated on the other two ``Unseen'' groups and ``All'' three groups, which are denoted by the blue bar and orange circle, respectively.}
	\label{fig:unseen}
	\vspace{-2mm}
\end{figure}

Considering the design of occlusion patterns as in Sec.~\ref{sec:proximity}, FROM is supposed to generalize well on unseen occlusions since it only depends on the location of occlusions instead of its type, which is empirically evaluated in Figure~\ref{fig:unseen}. All nine types of occluders are divided into three groups in alphabetical order: (1) [book, cup, eye mask]; (2) [eyeglasses, hand, face mask]; and (3) [phone, scarf, sunglasses]. Note that ``R$N$'' ($N \in [1,2,3]$) indicates the model is trained with the dataset occluded by $N_{th}$ group and evaluated on the other two unseen groups (\ie, ``Unseen'') and ``All'' (\ie, the dataset occluded by all three groups of occluders). Obviously, from Figure~\ref{fig:unseen}, the accuracy on ``Unseen'' dataset is basically not degraded, or even better than ``All'' for model ``R2'' and ``R3''. We conjecture this happened due to the size variance of occluders, \eg, the occluder phone in the third group usually has big size, resulting in ``R3'' generalizes even better on the other two ``Unseen'' groups with smaller size of occluders. Note that the accuracy of model trained and tested with all types occluders is $98.82\%$. In summary, thanks to the unique occlusion patterns design, FROM intrinsically has the excellent generalization ability to the unseen occlusions.

\begin{figure}[b]
	\centering
	\vspace{-2mm}
	\includegraphics[width=.9\linewidth]{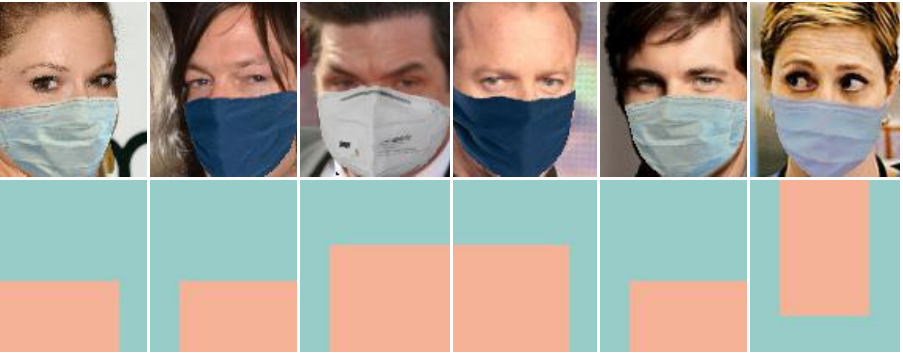}
	\caption{The first row contains masked face images generated by using FMA-3D~\cite{wang2021facex}, while their predicted occlusion patterns by FROM are included in the second row. The last column is the failure case.}
	\label{fig:mask}
	\vspace{-2mm}
\end{figure}

\begin{figure}[!htb]
	\centering
	\includegraphics[width=.85\linewidth]{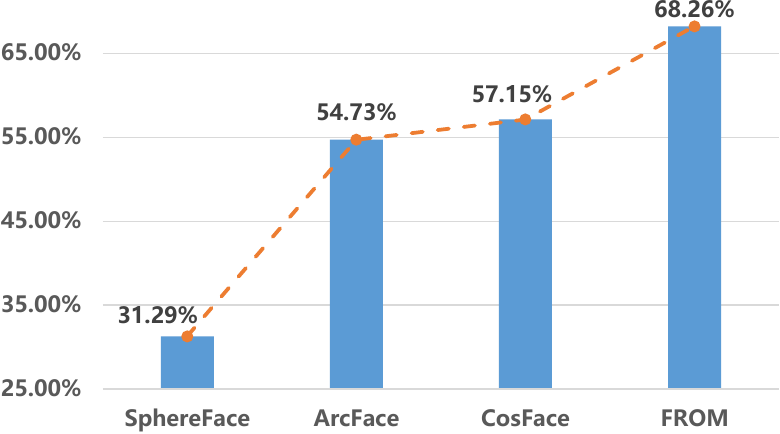}
	\caption{Rank-1 identification accuracy(\%) on Masked-MF1.}
	\label{fig:mask_result}
\end{figure}

\subsection{Application to Real-World Scenario}
To evaluate the effectiveness of FROM on real-world scenarios, we employ FROM on the masked face images, which are becoming common cases due to the recent world-wide COVID-19 pandemic. Following~\cite{wang2021facex}, we construct a photo-realistic masked facial dataset based on MegaFace by using FMA-3D~\cite{wang2021facex}, named Mask-MF1, which includes the masked probe images and remains the gallery images non-masked. The synthetic masked faces are high-fidelity and robust to different head pose, which are partially visualized in the first row of Figure~\ref{fig:mask}.

Several state-of-the-arts~\cite{liu2017sphereface,wang2018cosface,deng2019arcface} and FROM are tested on Mask-MF1 with rank-1 identification accuracy, as the results demonstrated in Figure~\ref{fig:mask_result}. Note that all the tested methods do not see the masked faces by FMA-3D during their training process. Not surprisingly, FROM achieves the best $68.26\%$ accuracy, significantly surpassing the $57.15\%$ and $54.73\%$ by CosFace and ArcFace, respectively. We also visualize some predicted occlusion patterns by FROM in the second row of Figure~\ref{fig:mask}. Those patterns reveal the location of occlusion and help FROM precisely remove the corrupted features to yield better recognition accuracy. The last column is the failure case which is expected given that FROM is not fed with the same kind of masked faces in the training. The above results and analysis suggest that FROM can be effectively and robustly applied on real-world scenarios.

\subsection{Benchmark on MFR2, LFW-SM and O\_LFW}
Both MFR2 and LFW-SM are proposed by MFRSA~\cite{anwar2020masked}. MFR2 is a small dataset of aligned masked faces with 53 identities of celebrities and politicians with a total of 269 images that are collected from the internet. A total of 848 image pairs from MFR2 (424 positive pairs, and 424 negative pairs) are considered for verification. LFW-SM contains images with the simulated mask (\eg, ``Surgical'', ``N95'', ``Cloth'' and ``Gas'') applied on LFW~\cite{huang2008labeled}. In addition, O\_LFW is constructed by BEL~\cite{shao2020biased} by employing 200 types of realistic occlusions to LFW, including 90 upper half occlusions, 70 lower half occlusions, 30 random occlusions and 10 large area occlusions. Verification between normal and occluded faces (N-O) and occluded face pairs (O-O) are used.

\begin{table}[!htb]
\center
\vspace{-2mm}
\caption{Face verification comparison ($\%$) on MFR2, LFW-SM and O\_LFW. N-O and O-O represent verification between normal and occluded faces and occluded face pairs of O\_LFW, respectively.}
\vspace{-2mm}
\label{table:sota}
\begin{tabular}{p{2cm}  p{2cm}<{\centering}  p{2cm}<{\centering}}
\hline
\noalign{\smallskip}
Method  & MFR2 & LFW-SM   \\
\noalign{\smallskip}
\hline
\noalign{\smallskip}
CosFace~\cite{wang2018cosface} & 93.99 & 96.22  \\
\noalign{\smallskip}
MFRSA~\cite{anwar2020masked} & 95.99 &	97.25    \\
\noalign{\smallskip}
FROM  & \textbf{96.22} &	\textbf{98.32}  \\
\noalign{\smallskip}
\hline
\noalign{\smallskip}
Method  &	N-O & O-O  \\
\noalign{\smallskip}
\hline
\noalign{\smallskip}
CosFace~\cite{wang2018cosface}  & 88.21 & 71.26 \\
\noalign{\smallskip}
BEL~\cite{shao2020biased}  & 88.12 & 75.10  \\
\noalign{\smallskip}
FROM  & \textbf{90.80} & \textbf{77.46}  \\
\noalign{\smallskip}
\hline
\end{tabular}
\end{table}
\vspace{2mm}

For the MFR2 and LFW-SM, we directly take the model trained on Occ-WebFace (without any finetuning) with $K=5, \lambda=1.0$ to evaluate, and the results are shown in the upper part of Table~\ref{table:sota}. Although our FROM is not fed with the same kind of masked face as in MFR2 and LFW-SM during training, it still surpasses MFRSA~\cite{anwar2020masked} and CosFace~\cite{wang2018cosface} on both MFR2 and LFW-SM, which is impressive considering FROM is not even specifically designed for masked face recognition. For O\_LFW, due to the images are with $112 \times 112$ size, we retrain FROM with $K=5, \lambda=1.0$ on the $112 \times 112$ version of Occ-WebFace (without any finetuning) while keeping other settings unchanged. From the results in the lower part of Table~\ref{table:sota}, we find that FROM consistently outperforms BEL~\cite{shao2020biased} and CosFace~\cite{wang2018cosface} on O\_LFW under both N-O and O-O protocols by a large margin. The above results forcefully support the superiority of FROM.

\subsection{Benchmark on Megaface Challenge 1}
In this experiment, we present benchmark results along with state-of-the-arts on the MegaFace Challenge 1 dataset (MF1), with and without occlusions. MF1, with one million distractors, is a much more challenging testing dataset compared to the LFW and AR face datasets. From Table~\ref{tab:mf_sota}, for clean face images, the Baseline achieves $73.64\%$ accuracy, while the proposed FROM obtains the higher $75.27\%$ accuracy. It reveals that our approach generalizes very well to clean face images, which can contribute to the special design for clean face image inside the \textbf{Occlusion Pattern Predictor}.

To test the performance on occluded face images, we evaluate the models on the ``Occ-MF1-1.0'' as described by Alg.~\ref{alg:construct} in Sec.~\ref{section:dataset}. According to Table~\ref{tab:mf_sota}, the Baseline only obtains $32.00\%$ accuracy, which is expected as it is not trained with occluded images. However, FROM achieve the best $60.84\%$ accuracy, which significantly improves about $28\%$ over the Baseline. This is because the feature masks produced by FROM mask out those corrupted feature elements accurately and preserve those clean and useful elements for later recognition. Since the ``Occ-MF1-1.0'' is newly constructed in this paper, other state-of-the-arts~\cite{liu2017sphereface,wang2018cosface,deng2019arcface} do not have the corresponding results. We thus evaluate those models with public resources on ``Occ-MF1-1.0'' by ourselves\footnotemark[1], which is denoted by $\dagger$ in Table~\ref{tab:mf_sota}. Apparently, ArcFace and CosFace achieve terrible accuracies $28.51\%$ and $31.31\%$, which are similar to the Baseline. SphereFace, without the powerful backbone (\ie, ResNet~\cite{he2016deep}), obtains the worst $17.46\%$ accuracy. Instead, our FROM remarkably outperforms them by nearly two times. It is also worth noting that the $56.34\%$ accuracy from PDSN~\cite{song2019occlusion} is the mean accuracy over 50 occluded areas, which, however, do not detailedly state in their paper. Thus, we can not obtain the fair-compared results under the same configuration. However, our $60.84\%$ accuracy is obtained on random location blocking which is more unconstrained and harder than 50 areas, making our results more convincing.

\begin{table}[!htb]
    \centering
    \caption{Face identification accuracy(\%) under protocol \textit{small} on MF1. Note that $\dagger$ represents the experimental results obtained by ourselves. \protect\footnotemark[1]}
    \label{tab:mf_sota}
\begin{tabular}{p{2.5cm}  p{1.5cm}<{\centering} p{2.0cm}<{\centering}}
\hline
\noalign{\smallskip}
Method  & MF1 \quad & Occ-MF1-1.0 \quad\\
\noalign{\smallskip}
\hline
\noalign{\smallskip}
CenterFace \cite{wen2016discriminative} & 65.49 & - \\
\noalign{\smallskip}
SphereFace \cite{liu2017sphereface} & 72.73 & 17.46$^{\dagger}$ \\
\noalign{\smallskip}
CosFace \cite{wang2018cosface} & 77.11 & 31.31$^{\dagger}$ \\
\noalign{\smallskip}
ArcFace \cite{deng2019arcface} & \textbf{77.50} & 28.51$^{\dagger}$ \\
\noalign{\smallskip}
PDSN  \cite{song2019occlusion}& 74.40 & 56.34 \\
\noalign{\smallskip}
\hline
\noalign{\smallskip}
Baseline  & 73.64  & 32.00 \\
\noalign{\smallskip}
FROM  & \textbf{75.27} & \textbf{60.84} \\
\hline
\end{tabular}
\end{table}
\footnotetext[1]{For SphereFace, we use the \href{https://github.com/clcarwin/sphereface_pytorch}{public code} and its pretrained model. For CosFace and ArcFace, we train the model with the corresponding loss by ourselves and obtain the experimental results.}

\subsection{Benchmark on AR Face Dataset}
Our method is evaluated on both ``Sunglasses'' and ``Scarf'' probe sets for the face identification task, and the results are shown in Table \ref{tab:ar_p12}. There are mainly two kinds of testing protocols in the existing literature \cite{weng2016robust,wan2017occlusion}. For protocol 1, it uses six images per person to form the gallery set, while only one image per person is used for protocol 2. The proposed FROM achieves $100\%$ accuracy on both ``Sunglasses'' and ``Scarf'' probe sets under protocol 1, which is a quite impressive performance given that we do not utilize any AR data to finetune our model. When facing the harder protocol 2, our FROM still obtains $100\%$ accuracy on both ``Sunglasses'' and ``Scarf'' probe sets, which significantly surpass the state-of-the-arts~\cite{liu2017sphereface,wang2018cosface,deng2019arcface}. It is also worth noting that closing the gap between good accuracies from~\cite{liu2017sphereface,wang2018cosface,deng2019arcface} and perfect accuracy achieved by FROM is not trivial. Results on both protocols have demonstrated that 
FROM consistently outperforms all previous state-of-the-arts.

\begin{table}[ht]
\centering
\caption{Face identification accuracy(\%) on the AR dataset under protocols 1, 2.  Note that ``Sg'' , ``Sf'' mean ``Sunglasses '' , ``Scarf '', respectively, and $\dagger$ represents the experimental results obtained by ourselves. \protect\footnotemark[1]}
\label{tab:ar_p12}
\begin{tabular}{p{1.8cm}  p{0.6cm}<{\centering} p{0.7cm}<{\centering} | p{1.8cm}  p{0.6cm}<{\centering} p{0.6cm}<{\centering}}
\cline{1-3}
\noalign{\smallskip}
Protocol 1  & Sg & Sf &   &   &  \\
\noalign{\smallskip}
\cline{1-3}
\noalign{\smallskip}
SRC \cite{wright2008robust}  & 87.00 & 59.50  &   &   & \\
\noalign{\smallskip}
NMR \cite{yang2016nuclear} & 96.90 & 73.50 &   &   & \\
\noalign{\smallskip}
\cline{4-6}
\noalign{\smallskip}
MLERPM \cite{weng2013robust}  & 98.00 & 97.00 & Protocol 2  & Sg & Sf \\
\noalign{\smallskip}
\cline{4-6}
\noalign{\smallskip}
SCF-PKR \cite{yang2013robust} & 95.65 & 98.00 & RPSM \cite{weng2016robust} & 84.84 & 90.16\\
\noalign{\smallskip}
RPSM \cite{weng2016robust} & 96.00 & 97.66 & Stringface \cite{chen2010recognizing} & 82.00 & 92.00 \\
\noalign{\smallskip}
MaskNet \cite{wan2017occlusion} & 90.90 & 96.70 & LMA \cite{mclaughlin2016largest} & 96.30 & 93.70  \\
\noalign{\smallskip}
PDSN \cite{song2019occlusion} & 99.72 & 100.0  & PDSN \cite{song2019occlusion}& 98.19 & 98.33 \\
\noalign{\smallskip}
\hline
\noalign{\smallskip}
SphereFace~\cite{liu2017sphereface}$^{\dagger}$ & 88.06 & 96.25 & SphereFace~\cite{liu2017sphereface}$^{\dagger}$  & 87.50 & 95.28\\
\noalign{\smallskip}
ArcFace~\cite{deng2019arcface}$^{\dagger}$ & 98.33 & 100.00 & ArcFace~\cite{deng2019arcface}$^{\dagger}$ & 95.00 & 99.72 \\
\noalign{\smallskip}
CosFace~\cite{wang2018cosface}$^{\dagger}$ & 99.44 & 100.00 & CosFace~\cite{wang2018cosface}$^{\dagger}$ & 98.06 & 100.00 \\
\noalign{\smallskip}
\hline
\noalign{\smallskip}
FROM  & \textbf{100.00} & \textbf{100.00} & FROM  & \textbf{100.00} & \textbf{100.00} \\
\noalign{\smallskip}
\hline
\end{tabular}
\end{table}
\section{Conclusions}
In this paper, we introduced FROM, a novel occlusion-robust face recognition method, learning to decode feature masks adaptively from the pyramid features that contain both local and global Information to remove corrupted face features caused by occlusions. Besides, FROM is simple yet powerful compared against the existing methods that either rely on external detectors to discover the occlusions, or employ shallow models that are less effective. At the training stage, FROM is trained in an end-to-end manner. At the testing time, it only requires a single forward pass to complete the mask generation and corrupted features removal, which makes FROM with an occlusion-robust characteristic ready to replace existing general face recognition systems nearly without an extra computational cost. The comprehensive experiments on LFW, Megaface Challenge 1, RMF2, AR datasets and other simulated occluded datasets demonstrate that our approach achieves superior accuracy for occluded face recognition, and generalizes very well on general face recognition. Furthermore, the evaluation of the generalization ability to unseen occlusions and the application to real-world scenarios (\eg, masked face recognition) verify the effectiveness and robustness of FROM.


\ifCLASSOPTIONcaptionsoff
  \newpage
\fi


%

\bibliographystyle{IEEEtran}
\bibliography{tpami2021}

\begin{thebibliography}{10}
\providecommand{\url}[1]{#1}
\csname url@samestyle\endcsname
\providecommand{\newblock}{\relax}
\providecommand{\bibinfo}[2]{#2}
\providecommand{\BIBentrySTDinterwordspacing}{\spaceskip=0pt\relax}
\providecommand{\BIBentryALTinterwordstretchfactor}{4}
\providecommand{\BIBentryALTinterwordspacing}{\spaceskip=\fontdimen2\font plus
\BIBentryALTinterwordstretchfactor\fontdimen3\font minus
  \fontdimen4\font\relax}
\providecommand{\BIBforeignlanguage}[2]{{%
\expandafter\ifx\csname l@#1\endcsname\relax
\typeout{** WARNING: IEEEtran.bst: No hyphenation pattern has been}%
\typeout{** loaded for the language `#1'. Using the pattern for}%
\typeout{** the default language instead.}%
\else
\language=\csname l@#1\endcsname
\fi
#2}}
\providecommand{\BIBdecl}{\relax}
\BIBdecl

\bibitem{deng2019arcface}
J.~Deng, J.~Guo, N.~Xue, and S.~Zafeiriou, ``Arcface: Additive angular margin
  loss for deep face recognition,'' in \emph{Proceedings of the IEEE Conference
  on Computer Vision and Pattern Recognition (CVPR)}, 2019, pp. 4690--4699.

\bibitem{hoffer2015deep}
E.~Hoffer and N.~Ailon, ``Deep metric learning using triplet network,'' in
  \emph{International Workshop on Similarity-Based Pattern Recognition}.\hskip
  1em plus 0.5em minus 0.4em\relax Springer, 2015, pp. 84--92.

\bibitem{liu2017sphereface}
W.~Liu, Y.~Wen, Z.~Yu, M.~Li, B.~Raj, and L.~Song, ``Sphereface: Deep
  hypersphere embedding for face recognition,'' in \emph{Proceedings of the
  IEEE conference on computer vision and pattern recognition (CVPR)}, 2017, pp.
  212--220.

\bibitem{wang2018cosface}
H.~Wang, Y.~Wang, Z.~Zhou, X.~Ji, D.~Gong, J.~Zhou, Z.~Li, and W.~Liu,
  ``Cosface: Large margin cosine loss for deep face recognition,'' in
  \emph{Proceedings of the IEEE Conference on Computer Vision and Pattern
  Recognition (CVPR)}, 2018, pp. 5265--5274.

\bibitem{wen2016discriminative}
Y.~Wen, K.~Zhang, Z.~Li, and Y.~Qiao, ``A discriminative feature learning
  approach for deep face recognition,'' in \emph{European Conference on
  Computer Vision (ECCV)}.\hskip 1em plus 0.5em minus 0.4em\relax Springer,
  2016, pp. 499--515.

\bibitem{he2016deep}
K.~He, X.~Zhang, S.~Ren, and J.~Sun, ``Deep residual learning for image
  recognition,'' in \emph{Proceedings of the IEEE conference on computer vision
  and pattern recognition (CVPR)}, 2016, pp. 770--778.

\bibitem{huang2017densely}
G.~Huang, Z.~Liu, L.~Van Der~Maaten, and K.~Q. Weinberger, ``Densely connected
  convolutional networks,'' in \emph{Proceedings of the IEEE Conference on
  Computer Vision and Pattern Recognition (CVPR)}, 2017, pp. 4700--4708.

\bibitem{krizhevsky2017imagenet}
A.~Krizhevsky, I.~Sutskever, and G.~E. Hinton, ``Imagenet classification with
  deep convolutional neural networks,'' \emph{Communications of the ACM},
  vol.~60, no.~6, pp. 84--90, 2017.

\bibitem{simonyan2014very}
K.~Simonyan and A.~Zisserman, ``Very deep convolutional networks for
  large-scale image recognition,'' \emph{arXiv preprint arXiv:1409.1556}, 2014.

\bibitem{tan2019efficientnet}
M.~Tan and Q.~V. Le, ``Efficientnet: Rethinking model scaling for convolutional
  neural networks,'' \emph{arXiv preprint arXiv:1905.11946}, 2019.

\bibitem{guo2016ms}
Y.~Guo, L.~Zhang, Y.~Hu, X.~He, and J.~Gao, ``Ms-celeb-1m: A dataset and
  benchmark for large-scale face recognition,'' in \emph{European Conference on
  Computer Vision (ECCV)}.\hskip 1em plus 0.5em minus 0.4em\relax Springer,
  2016, pp. 87--102.

\bibitem{huang2008labeled}
G.~B. Huang, M.~Mattar, T.~Berg, and E.~Learned-Miller, ``Labeled faces in the
  wild: A database forstudying face recognition in unconstrained
  environments,'' 2008.

\bibitem{kemelmacher2016megaface}
I.~Kemelmacher-Shlizerman, S.~M. Seitz, D.~Miller, and E.~Brossard, ``The
  megaface benchmark: 1 million faces for recognition at scale,'' in
  \emph{Proceedings of the IEEE conference on computer vision and pattern
  recognition (CVPR)}, 2016, pp. 4873--4882.

\bibitem{mehdipour2016comprehensive}
M.~Mehdipour~Ghazi and H.~Kemal~Ekenel, ``A comprehensive analysis of deep
  learning based representation for face recognition,'' in \emph{Proceedings of
  the IEEE conference on computer vision and pattern recognition workshops
  (CVPR Workshops)}, 2016, pp. 34--41.

\bibitem{yi2014learning}
D.~Yi, Z.~Lei, S.~Liao, and S.~Z. Li, ``Learning face representation from
  scratch,'' \emph{arXiv preprint arXiv:1411.7923}, 2014.

\bibitem{zhao2017robust}
F.~Zhao, J.~Feng, J.~Zhao, W.~Yang, and S.~Yan, ``Robust lstm-autoencoders for
  face de-occlusion in the wild,'' \emph{IEEE Transactions on Image
  Processing}, vol.~27, no.~2, pp. 778--790, 2017.

\bibitem{song2019occlusion}
L.~Song, D.~Gong, Z.~Li, C.~Liu, and W.~Liu, ``Occlusion robust face
  recognition based on mask learning with pairwise differential siamese
  network,'' in \emph{Proceedings of IEEE International Conference on Computer
  Vision (ICCV)}, 2019, pp. 773--782.

\bibitem{chopra2005learning}
S.~Chopra, R.~Hadsell, and Y.~LeCun, ``Learning a similarity metric
  discriminatively, with application to face verification,'' in
  \emph{Proceedings of the IEEE Conference on Computer Vision and Pattern
  Recognition (CVPR)}, vol.~1.\hskip 1em plus 0.5em minus 0.4em\relax IEEE,
  2005, pp. 539--546.

\bibitem{hadsell2006dimensionality}
R.~Hadsell, S.~Chopra, and Y.~LeCun, ``Dimensionality reduction by learning an
  invariant mapping,'' in \emph{Proceedings of the IEEE Conference on Computer
  Vision and Pattern Recognition (CVPR)}, vol.~2.\hskip 1em plus 0.5em minus
  0.4em\relax IEEE, 2006, pp. 1735--1742.

\bibitem{liu2016large}
W.~Liu, Y.~Wen, Z.~Yu, and M.~Yang, ``Large-margin softmax loss for
  convolutional neural networks.'' in \emph{International Conference on Machine
  Learning (ICML)}, vol.~2, no.~3, 2016, p.~7.

\bibitem{duan2019uniformface}
Y.~Duan, J.~Lu, and J.~Zhou, ``Uniformface: Learning deep equidistributed
  representation for face recognition,'' in \emph{Proceedings of the IEEE
  Conference on Computer Vision and Pattern Recognition}, 2019, pp. 3415--3424.

\bibitem{wright2008robust}
J.~Wright, A.~Y. Yang, A.~Ganesh, S.~S. Sastry, and Y.~Ma, ``Robust face
  recognition via sparse representation,'' \emph{IEEE transactions on pattern
  analysis and machine intelligence (PAMI)}, vol.~31, no.~2, pp. 210--227,
  2008.

\bibitem{zhou2009face}
Z.~Zhou, A.~Wagner, H.~Mobahi, J.~Wright, and Y.~Ma, ``Face recognition with
  contiguous occlusion using markov random fields,'' in \emph{Proceedings of
  IEEE International Conference on Computer Vision (ICCV)}.\hskip 1em plus
  0.5em minus 0.4em\relax IEEE, 2009, pp. 1050--1057.

\bibitem{li2005nonparametric}
Z.~Li, W.~Liu, D.~Lin, and X.~Tang, ``Nonparametric subspace analysis for face
  recognition,'' in \emph{Proceedings of the IEEE conference on computer vision
  and pattern recognition (CVPR)}, vol.~2.\hskip 1em plus 0.5em minus
  0.4em\relax IEEE, 2005, pp. 961--966.

\bibitem{wan2017occlusion}
W.~Wan and J.~Chen, ``Occlusion robust face recognition based on mask
  learning,'' in \emph{IEEE International Conference on Image Processing
  (ICIP)}.\hskip 1em plus 0.5em minus 0.4em\relax IEEE, 2017, pp. 3795--3799.

\bibitem{yang2011robust}
M.~Yang, L.~Zhang, J.~Yang, and D.~Zhang, ``Robust sparse coding for face
  recognition,'' in \emph{Proceedings of the IEEE conference on computer vision
  and pattern recognition (CVPR)}.\hskip 1em plus 0.5em minus 0.4em\relax IEEE,
  2011, pp. 625--632.

\bibitem{he2011regularized}
R.~He, W.-S. Zheng, B.-G. Hu, and X.-W. Kong, ``A regularized correntropy
  framework for robust pattern recognition,'' \emph{Neural computation},
  vol.~23, no.~8, pp. 2074--2100, 2011.

\bibitem{li2013structured}
X.-X. Li, D.-Q. Dai, X.-F. Zhang, and C.-X. Ren, ``Structured sparse error
  coding for face recognition with occlusion,'' \emph{IEEE transactions on
  image processing}, vol.~22, no.~5, pp. 1889--1900, 2013.

\bibitem{min2011improving}
R.~Min, A.~Hadid, and J.-L. Dugelay, ``Improving the recognition of faces
  occluded by facial accessories,'' in \emph{Face and Gesture}.\hskip 1em plus
  0.5em minus 0.4em\relax IEEE, 2011, pp. 442--447.

\bibitem{oh2008occlusion}
H.~J. Oh, K.~M. Lee, and S.~U. Lee, ``Occlusion invariant face recognition
  using selective local non-negative matrix factorization basis images,''
  \emph{Image and Vision computing}, vol.~26, no.~11, pp. 1515--1523, 2008.

\bibitem{lin2017feature}
T.-Y. Lin, P.~Doll{\'a}r, R.~Girshick, K.~He, B.~Hariharan, and S.~Belongie,
  ``Feature pyramid networks for object detection,'' in \emph{Proceedings of
  the IEEE conference on computer vision and pattern recognition}, 2017, pp.
  2117--2125.

\bibitem{long2015fully}
J.~Long, E.~Shelhamer, and T.~Darrell, ``Fully convolutional networks for
  semantic segmentation,'' in \emph{Proceedings of the IEEE conference on
  computer vision and pattern recognition (CVPR)}, 2015, pp. 3431--3440.

\bibitem{ng2014data}
H.-W. Ng and S.~Winkler, ``A data-driven approach to cleaning large face
  datasets,'' in \emph{IEEE international conference on image processing
  (ICIP)}.\hskip 1em plus 0.5em minus 0.4em\relax IEEE, 2014, pp. 343--347.

\bibitem{martinez1998ar}
A.~M. Martinez, ``The ar face database,'' \emph{CVC Technical Report24}, 1998.

\bibitem{zhang2016joint}
K.~Zhang, Z.~Zhang, Z.~Li, and Y.~Qiao, ``Joint face detection and alignment
  using multitask cascaded convolutional networks,'' \emph{IEEE Signal
  Processing Letters}, vol.~23, no.~10, pp. 1499--1503, 2016.

\bibitem{van2008visualizing}
L.~Van~der Maaten and G.~Hinton, ``Visualizing data using t-sne.''
  \emph{Journal of machine learning research}, vol.~9, no.~11, 2008.

\bibitem{wang2021facex}
J.~Wang, Y.~Liu, Y.~Hu, H.~Shi, and T.~Mei, ``Facex-zoo: A pytorh toolbox for
  face recognition,'' \emph{arXiv preprint arXiv:2101.04407}, 2021.

\bibitem{anwar2020masked}
A.~Anwar and A.~Raychowdhury, ``Masked face recognition for secure
  authentication,'' 2020.

\bibitem{shao2020biased}
C.~Shao, J.~Huo, L.~Qi, Z.-H. Feng, W.~Li, C.~Dong, and Y.~Gao, ``Biased
  feature learning for occlusion invariant face recognition.'' in \emph{IJCAI},
  2020, pp. 666--672.

\bibitem{weng2016robust}
R.~Weng, J.~Lu, and Y.-P. Tan, ``Robust point set matching for partial face
  recognition,'' \emph{IEEE transactions on image processing}, vol.~25, no.~3,
  pp. 1163--1176, 2016.

\bibitem{yang2016nuclear}
J.~Yang, L.~Luo, J.~Qian, Y.~Tai, F.~Zhang, and Y.~Xu, ``Nuclear norm based
  matrix regression with applications to face recognition with occlusion and
  illumination changes,'' \emph{IEEE transactions on pattern analysis and
  machine intelligence (PAMI)}, vol.~39, no.~1, pp. 156--171, 2016.

\bibitem{weng2013robust}
R.~Weng, J.~Lu, J.~Hu, G.~Yang, and Y.-P. Tan, ``Robust feature set matching
  for partial face recognition,'' in \emph{Proceedings of the IEEE
  International Conference on Computer Vision (ICCV)}, 2013, pp. 601--608.

\bibitem{yang2013robust}
M.~Yang, L.~Zhang, S.~C.-K. Shiu, and D.~Zhang, ``Robust kernel representation
  with statistical local features for face recognition,'' \emph{IEEE
  transactions on neural networks and learning systems}, vol.~24, no.~6, pp.
  900--912, 2013.

\bibitem{chen2010recognizing}
W.~Chen and Y.~Gao, ``Recognizing partially occluded faces from a single sample
  per class using string-based matching,'' in \emph{European Conference on
  Computer Vision (ECCV)}.\hskip 1em plus 0.5em minus 0.4em\relax Springer,
  2010, pp. 496--509.

\bibitem{mclaughlin2016largest}
N.~McLaughlin, J.~Ming, and D.~Crookes, ``Largest matching areas for
  illumination and occlusion robust face recognition,'' \emph{IEEE transactions
  on cybernetics}, vol.~47, no.~3, pp. 796--808, 2016.

\end{thebibliography}

\begin{IEEEbiography}[{\includegraphics[width=1in,height=1.25in,clip,keepaspectratio]{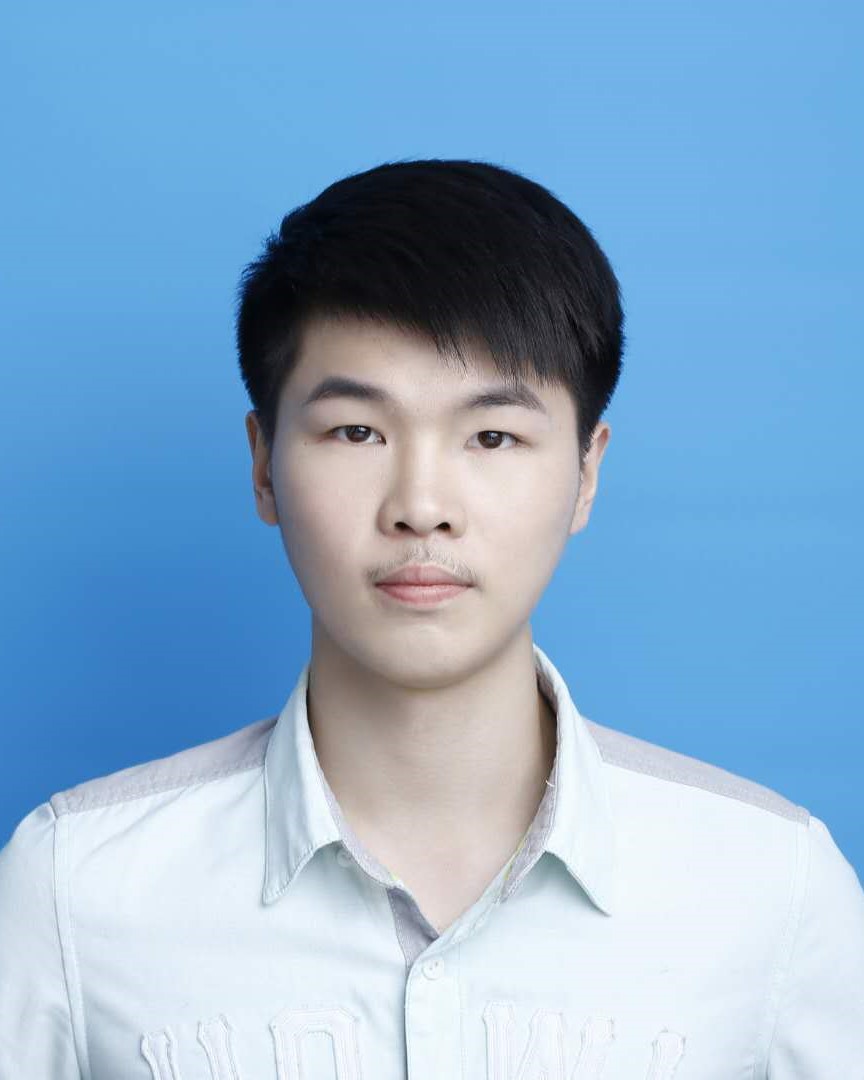}}]{Haibo Qiu}
received the B.Eng. degree in the Department of Electronic Engineering and Information Science from the University of Science and Technology of China, China, 2018. His research interests are in computer vision and deep learning. Currently, he is a PhD candidate at the School of Computer Science in the University of Sydney, supervised by Prof. Dacheng Tao.
\end{IEEEbiography}

\begin{IEEEbiography}[{\includegraphics[width=1in,height=1.25in,clip,keepaspectratio]{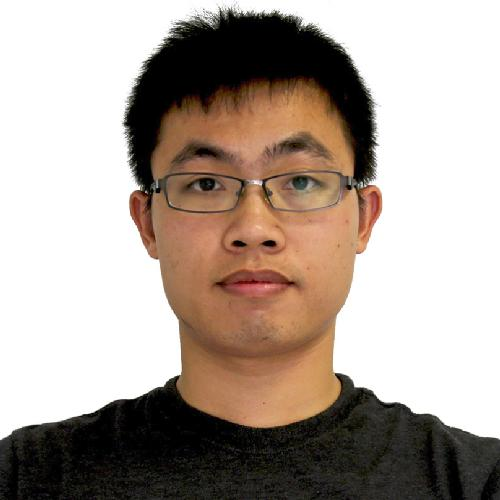}}]{Dihong Gong}
received the Ph.D. degree of computer science from the University of Florida, in 2018. He then joined the Tencent AI Lab as a senior research scientist, with research interest primarily focused on face related technologies, including face detection, recognition, liveness examination.
\end{IEEEbiography}
\vfill

\begin{IEEEbiography}[{\includegraphics[width=1in,height=1.25in,clip,keepaspectratio]{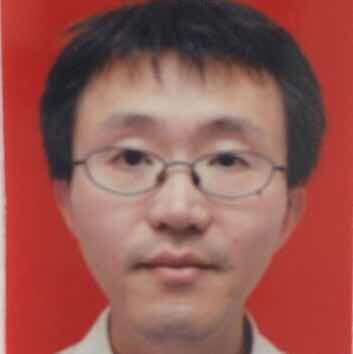}}]{Zhifeng Li}
(M'06-SM'11) is currently a top-tier principal research scientist with Tencent. Before joining Tencent, he was a full professor with the Shenzhen Institutes of Advanced Technology, Chinese Academy of Sciences. His research interests include deep learning, computer vision and pattern recognition, and face detection and recognition. He was one of the 2020 Most Cited Chinese Researchers (Elsevier-Scopus) in computer science and technology. He is currently serving on the Editorial Boards of Neurocomputing and IEEE Transactions on Circuits and Systems for Video Technology. He is a fellow of British Computer Society (FBCS).
\end{IEEEbiography}

\vfill
\enlargethispage{-2.5in}

\begin{IEEEbiography}[{\includegraphics[width=1in,height=1.25in,clip,keepaspectratio]{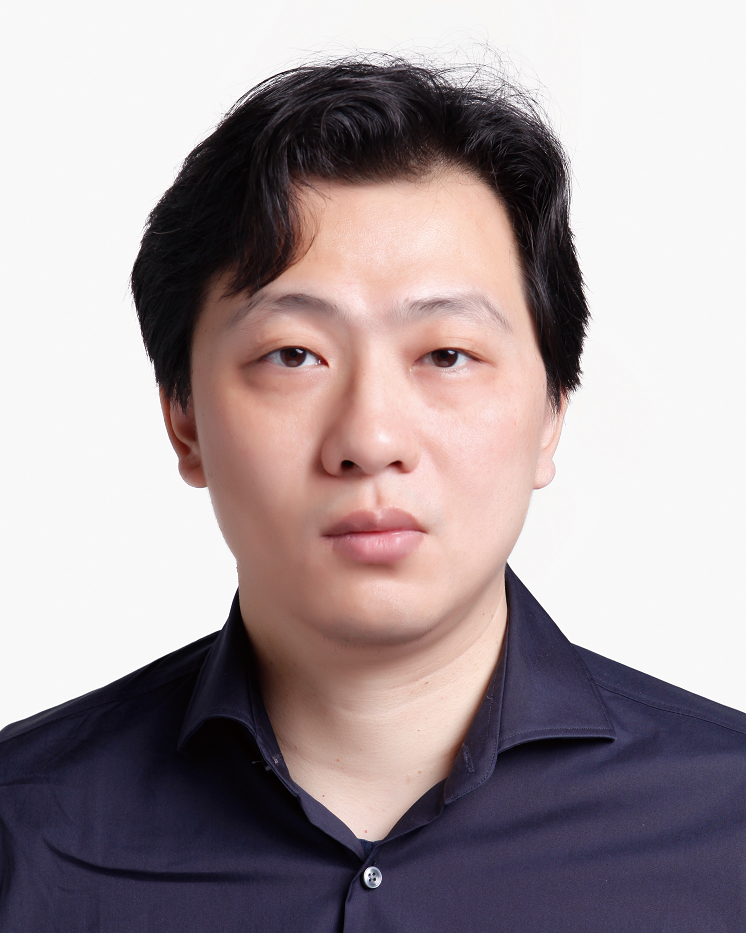}}]{Wei Liu}
(M'14-SM'19) is currently a Distinguished Scientist of Tencent, China and a director of Computer Vision Center at Tencent AI Lab.  Prior to that, he has been a research staff member of IBM T. J. Watson Research Center, Yorktown Heights, NY, USA from 2012 to 2015. Dr. Liu has long been devoted to research and development in the fields of machine learning, computer vision, pattern recognition, information retrieval, big data, etc. Dr. Liu currently serves on the editorial boards of IEEE Transactions on Pattern Analysis and Machine Intelligence, IEEE Transactions on Neural Networks and Learning Systems, IEEE Transactions on Circuits and Systems for Video Technology, Pattern Recognition, etc. He is a Fellow of the International Association for Pattern Recognition (IAPR) and an Elected Member of the International Statistical Institute (ISI). 
\end{IEEEbiography}

\begin{IEEEbiography}[{\includegraphics[width=1in,height=1.25in,clip,keepaspectratio]{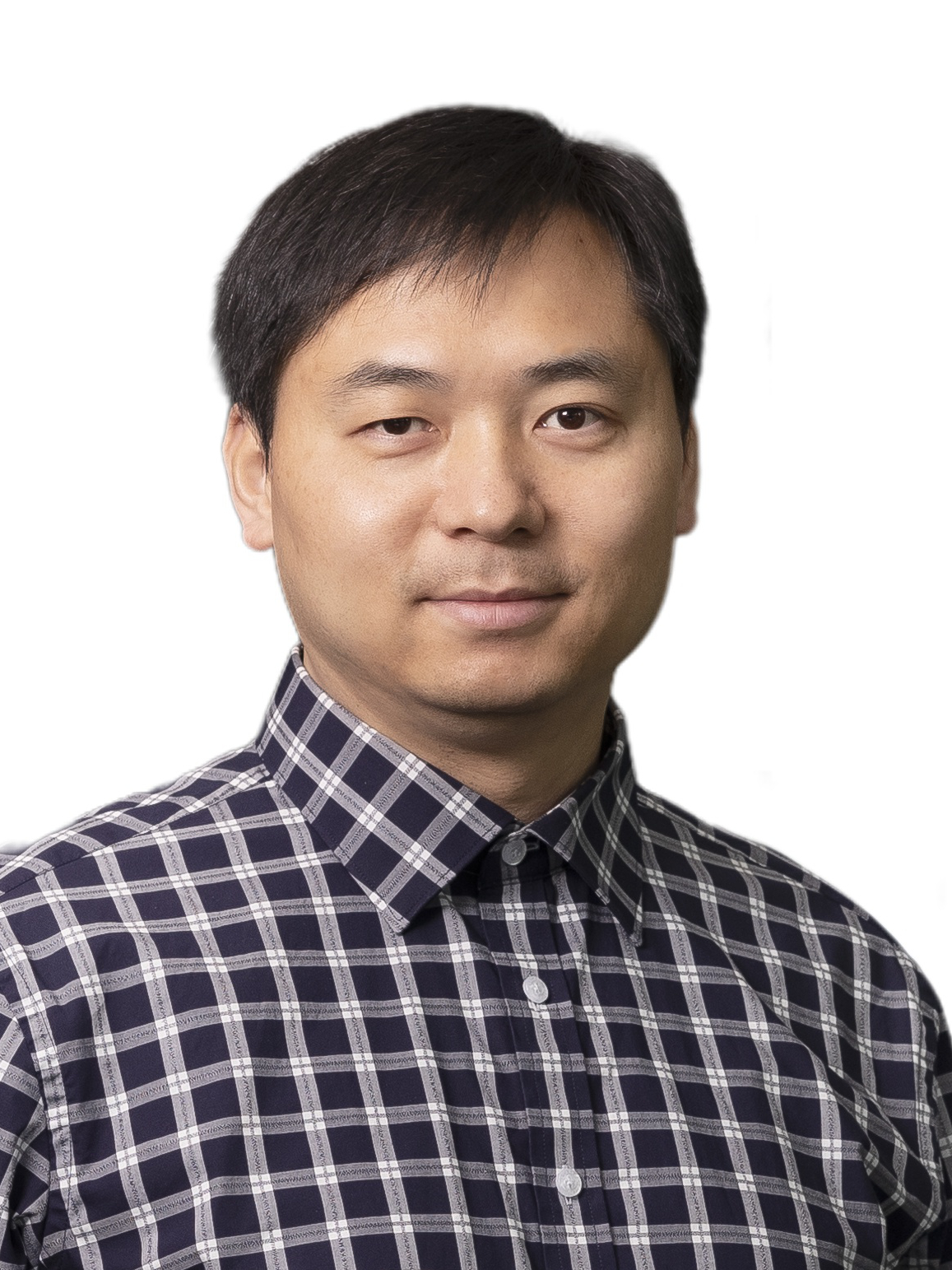}}]{Dacheng Tao}
(F’15) is currently the president of the JD Explore Academy and a senior vice president of JD.com. He mainly applies statistics and mathematics to artificial intelligence and data science, and his research is detailed in one monograph and over 200 publications in prestigious journals and proceedings at leading conferences. He received the 2015/2020 Australian Eureka Prize, the 2018 IEEE ICDM Research Contributions Award, and the 2021 IEEE Computer Society McCluskey Technical Achievement Award. He is a fellow of the Australian Academy of Science, AAAS, ACM and IEEE.
\end{IEEEbiography}
\vfill


%

\end{document}